\documentclass[10pt,twocolumn,letterpaper]{article}

\usepackage[pagenumbers]{cvpr} %

\usepackage{cuted} %

\usepackage{currfile} %

\usepackage{caption} %
\usepackage{longtable,array} %
\usepackage{multirow} %

\usepackage{xspace}
\newcommand{\vlens}{\textsc{V-lens}\xspace}
\newcommand{\vlensicon}{\raisebox{-0.15em}{\includegraphics[height=0.9em]{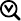}}\xspace}
\newcommand{\bagel}{BAGEL\xspace}
\newcommand{\showo}{Show-o2\xspace}
\newcommand{\sensenova}{SenseNova-U1\xspace}
\newcommand{\liquid}{Liquid\xspace}
\newcommand{\januspro}{Janus-Pro\xspace}
\newcommand{\hunyuan}{HunyuanImage-3.0\xspace}

\newif\iffullres
\fullrestrue
\iffullres\graphicspath{{figures/out/}}\else\graphicspath{{figures/out/low/}{figures/out/}}\fi
\newcommand{\suppref}[1]{Supp.~\labelcref{#1}}
\newcommand{\storydiff}[1]{{\color{black!55}#1}}                 %

\newif\ifwithsupp
\withsupptrue
\ifwithsupp\else\usepackage{xr-hyper}\fi

\definecolor{cvprblue}{rgb}{0.21,0.49,0.74}
\usepackage[pagebackref,breaklinks,colorlinks,allcolors=cvprblue]{hyperref}
\usepackage[capitalize]{cleveref}



\def\confName{CVPR}
\def\confYear{2027}

\title{The Elephant in the Question: \\
Unified Models Picture the Concepts Behind Their Answers}

\author{%
  Tobias Braun$^{1,2}$ \qquad Marcus Rohrbach$^{1}$ \qquad Lorenzo Cavallaro$^{3,2}$\\[4pt]
  {\small $^{1}$Technische Universit\"at Darmstadt \quad $^{2}$University College London \quad
   $^{3}$Mohamed bin Zayed University of Artificial Intelligence}
}

\begin{document}
\maketitle
\begin{strip}
\vspace{-1.2cm}
  \centering
  \includegraphics[width=\textwidth]{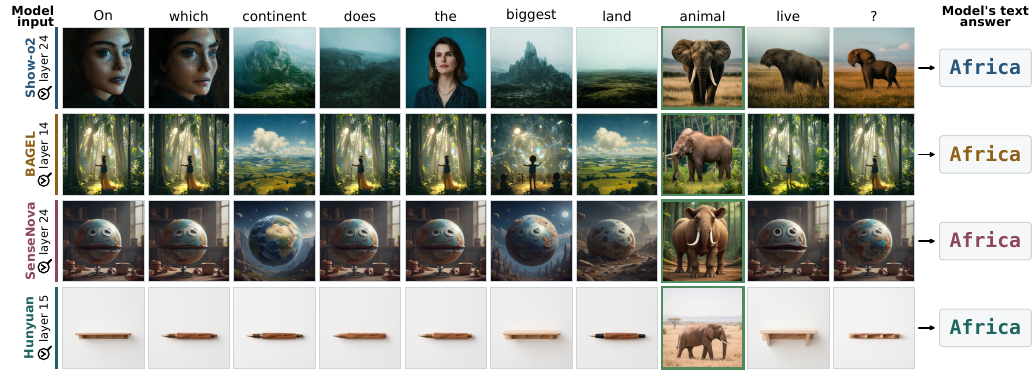}
  \captionof{figure}{All four unified vision-language models correctly answer the question with ``Africa'' without mentioning the ``elephant'' explicitly. The \vlens{}\,\vlensicon{} reveals the elephant as an intermediate concept in the computation leading to their answer. It uses each model's own image generator to visualize hidden states at each question token and surface reasoning concepts that remain unspoken in the text.}
  \label{fig:teaser}
\end{strip}

\begin{abstract}
Visualizing intermediate representations could help us inspect the reasoning steps underlying a model output. Unified multimodal models offer this opportunity by combining understanding and generation in one architecture. Their image generators can access representations formed while processing text. We introduce the \vlens{}, which uses this connection to visualize a chosen token's hidden state without training a decoder. Across four diffusion-based and two purely autoregressive unified models, the resulting images reveal key intermediate concepts that connect a question to its answer even when the concepts are never named explicitly. Controlled tests show how this content changes with context, token position, and depth. %
We then introduce lens tuning, which supervises selected hidden states with pictures through the \vlens{}. Across \bagel{}, \sensenova{}, and \hunyuan{}, the rate of correct visualizations of pronoun referents rises from 2--11\% to 86--94\%. Surprisingly, such visual supervision also improves text answers without supervising the answer tokens directly. Code is available at {\urlstyle{same}\url{https://github.com/multimodal-ai-lab/V-lens}}. %
\end{abstract}

\section{Introduction}
\label{sec:intro}

To answer \emph{On which continent does the biggest land animal live?}, a model must connect the question to an elephant and the elephant to Africa. The models in \cref{fig:teaser} correctly answer \emph{Africa} without ever generating the word \emph{elephant}. During the generation of their text answers, the models must therefore internally represent the concept of an elephant before shifting their representation toward the output token Africa. While prior work has developed methods to \emph{verbalize} such intermediate reasoning concepts \citep{frasertaliente2026nla,gurnee2026workspace}, we aim for a complementary route. Motivated by human reasoning, which can draw on mental imagery~\citep{pearson2019imagination,hegarty2004mechanical} to solve problems, we aim to \emph{visualize} them using the model's own means of visualization. Specifically, we ask:
\par\vspace{0.3cm}{\centering\itshape Can intermediate concepts be made visible within a\\ model's computation without training a separate decoder?\par}\vspace{0.3cm}\noindent We find that the model's own image generator can indeed visualize these concepts from text token representations in its residual stream. These visualizations offer a new way to inspect and supervise content that the text response omits.

Unified multimodal models make this possible through their architecture. They combine understanding and generation across text and images in one system, allowing the image generator to read representations computed while understanding text. Building on this connection, we introduce the \vlens{}. Transformer layers maintain contextualized token representations known as \emph{hidden states}~\citep{vaswani2017attention}. We center and rescale a selected hidden state to account for differences in the average and spread of states across layers. We then insert it into a minimal drawing prompt and let the model’s own generator draw an image. This requires no decoder training. Prior work makes a model's intermediate states interpretable as text tokens~\citep{nostalgebraist2020logitlens,belrose2023tunedlens,gurnee2026workspace}. Our approach visualizes them as images and traces how these visualizations evolve during text generation. As the \vlens{} uses the model's own image generator, it is not restricted to one specific generation paradigm. We show that our approach works for models that process text
and image within one shared transformer \citep{xie2025showo2, hunyuan2025image3}, models that also use a separate set of weights during image generation \citep{deng2025bagel, diao2026sensenova}, and purely autoregressive generators that draw token by token \citep{chen2025januspro, wu2024liquid}.

To assess the level of detail captured by the \vlens{}, we vary the context around specific text tokens, such as object references and words with two meanings. We find that the same token can yield different visualizations depending on its context, most consistently for words with two meanings.
An analysis across layers reveals that unnamed objects can be visualized at intermediate depths and fade toward the final layers in most models.
The link from a hidden state to its \vlens{} picture can also be trained. \bagel{}~\citep{deng2025bagel}, \sensenova{}~\citep{diao2026sensenova}, and \hunyuan{}~\citep{hunyuan2025image3} often answer reference questions correctly while the visualization of the referent shows the incorrect object. We add small adapters to the \emph{text} layers and supervise the hidden state with self-generated \emph{images} through the \vlens{}. On held-out texts, correct visualizations rise from 2\% to 86\% in \bagel{}, 11\% to 94\% in \sensenova{}, and 6\% to 86\% in \hunyuan{}. We call this procedure \textit{lens tuning} and also test it with questions in which an event changes an object's state. Despite purely visual supervision, lens tuning improves text answers on supervised stories in \showo{}~\citep{xie2025showo2} and \hunyuan{}. Finally, transferring a hidden state from a lens-tuned model to the original model carries part of the correction, showing that the states we visualize are directly connected to the model's text answers.

Our work makes the following three contributions:
\begin{itemize}
    \item The \vlens{}, which visualizes intermediate hidden states of text tokens without training a separate decoder.
    \item Controlled tests in six unified models showing that visualizations follow a token's context and, in five of them, reveal unnamed concepts at specific tokens and depths.
    \item Lens tuning that uses only visual supervision through the  \vlens{} to improve visualizations \textit{and} text answers. 
\end{itemize}

\section{Related work}
\label{sec:related}

\paragraph{Reading hidden states.}

Prior work makes neural computations more interpretable by visualizing learned features~\citep{olah2017feature} and identifying computational subgraphs (circuits) that implement specific behaviors~\citep{olah2020zoom}. Sparse autoencoders reconstruct activations through combinations of learned features that are often easier to interpret than individual neurons~\citep{huben2024sparse}. Natural language autoencoders learn to reconstruct activations through text descriptions~\citep{frasertaliente2026nla}, while activation oracles learn to answer questions about them~\citep{karvonen2026activation}.
Hidden-state lenses also aim to make intermediate computations interpretable for humans. The logit lens reads earlier states through the model's final word prediction layer~\citep{nostalgebraist2020logitlens}, an approach also used to analyze visual tokens in vision-language models~\citep{neo2025interpreting}. Because states at different layers can use different representations, the tuned lens first learns a linear map and offset for each layer to match the model's final token predictions~\citep{belrose2023tunedlens}. The Jacobian lens, or J-lens, uses averaged derivatives of the model's computation to carry a state toward the final layer before reading it through the vocabulary~\citep{gurnee2026workspace}. Other methods use hidden states to estimate uncertainty~\citep{braun2026uspace}, assess truthfulness~\citep{azaria2023internal}, and detect harmful interactions~\citep{mckenzie2025detecting}.  This work's \vlens{} translates hidden states into images as an additional pathway for analysis and model improvement.

\paragraph{Interpreting hidden states through new prompts.}
Patchscopes and SelfIE use a model's ability to generate text to explain its hidden states. They place a state into a new prompt to describe its content~\citep{ghandeharioun2024patchscopes,chen2024selfie}, though decoding can favor the model's own prior over the patched context~\citep{gong2026faithful}. The \vlens{} follows the Patchscopes construction with an image as the response, using the model's own image generator. Activation patching also replaces states, but measures changes in the original task to test their causal role~\citep{meng2022locating}. Lindsey injects concept representations into a model's activations to test whether it notices and names them~\citep{lindsey2026introspection}. %
Work on multi-step reasoning asks whether a model identifies an unnamed concept in time to use it when answering~\citep{yang2024multihop,biran2024hopping}. Our questions use their two-step structure to visualize connecting objects from hidden states.

\paragraph{Reading text representations as pictures.}
Diffusion Lens is the closest prior method~\citep{toker2024diffusion}. It decodes intermediate text-encoder representations through a separate diffusion model to trace the image prompt across layers. Diffusion Steering Lens visualizes vision transformer submodule outputs through diffusion~\citep{takatsuki2025diffusion}. The \vlens{} extends this idea to unified models that generate images \emph{and} text without depending on an external text-encoder. It enables the visualization of a selected hidden state within questions, stories, or their own replies rather than mere image generation prompts. We center and rescale each coordinate of a chosen state before inserting it into a drawing prompt for the same model. This reads text computations through denoising or autoregressive image generation, complementing work that uses visual grounding or imagined pictures to aid language understanding and reasoning~\citep{tan2020vokenization,yang2022zlavi,lu2022imagination,yang2025mirage,singhi2026reasoning}.

\paragraph{Unified models.}
\citet{cho2020xlxmert} were among the first to explore unifying multimodal understanding and generation within one transformer framework. Subsequent models differ in parameter sharing and image generation objectives. The fully unified models \showo{} and \hunyuan{} process text and image tokens through a shared transformer backbone~\citep{xie2025showo2,hunyuan2025image3}, while \bagel{} and \sensenova{} use separate transformer blocks connected through attention~\citep{deng2025bagel,diao2026sensenova}. These models generate images through diffusion or flow matching. Autoregressive alternatives such as \liquid{} and \januspro{} instead predict discrete image tokens sequentially~\citep{wu2024liquid,chen2025januspro}. We show that the \vlens{} works for these different architectures and generation paradigms.

\section{The \vlens{}}
\label{sec:method}

\begin{figure}[t]
    \vspace{-0.35cm}
    \centering
    \includegraphics[width=\columnwidth]{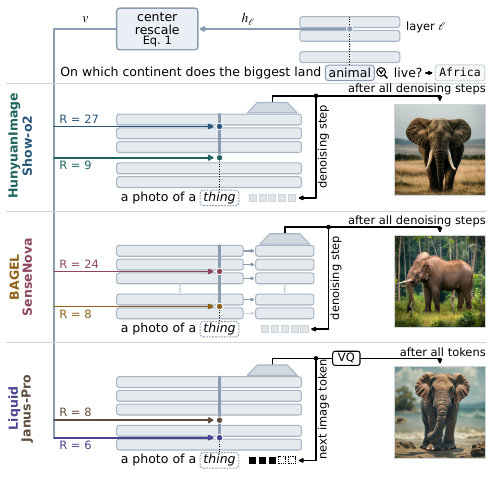}
    \caption{The \vlens{} visualizes a hidden state through the model's own image generator. Top, we take the state entering layer $\ell$ at the token \emph{animal} and center and rescale it with \cref{eq:transport}. The resulting state replaces \emph{thing} in the drawing request as it enters the insertion layer $R$. \hunyuan{} and \showo{} process text and image with one shared transformer. \bagel{} and \sensenova{} draw with a second copy of the weights that reads the request's keys and values at every layer. \liquid{} and \januspro{} write the image one token at a time after attending prior text and image tokens.
    }
    \vspace{-0.18cm}
    \label{fig:method}
\end{figure}

In their responses, modern language models may use concepts that they do not mention explicitly. To surface such concepts visually, the \vlens{} centers and rescales a text token's intermediate hidden state before inserting it in a dedicated drawing request. We call the layer where we take the state the \emph{extraction layer} $\ell$, and the layer where we place it the \emph{insertion layer} $R$. We fix $R$ per model and vary $\ell$ to read a token at different depths. \Cref{fig:method} shows how a state reaches the image generator in three kinds of unified models. We then describe \emph{lens tuning}, which trains hidden states on images via the \vlens{} to improve their visualizations, and examine how this affects the model's text answers.

\paragraph{The insertion layer.}
\label{sec:insertion-layer}
We calibrate the insertion layer to find the latest point where the model can still draw recognizable objects from hidden states. We take 12 common, visually distinct nouns from the concept lists of \citet{udandarao2024zeroshot}, which CLIP~\citep{radford2021clip} recognizes reliably~\citep{parashar2024neglected,udandarao2024zeroshot}. Each word's state from an ordinary sentence replaces \emph{thing} in \emph{a photo of a thing} at the same layer. We count how often CLIP ranks the intended word first among 100 concept names and choose the latest tested layer with the highest count. This gives $R=8$ of 28 for \bagel{}, $R=27$ of 28 for \showo{}, $R=24$ of 42 for \sensenova{}, $R=9$ of 32 for \hunyuan{}, $R=6$ of 28 for \liquid{}, and $R=8$ of 30 for \januspro{}, each recovering 12/12 words. All calibration and sampling settings are in \suppref{supp:implementation,supp:autoregressive}.

\paragraph{Carrying a hidden state between layers.}
\label{sec:transport}
The \vlens{} matches the state's center and spread to the insertion layer so that the model can draw from states at different depths. Let $h_{\ell,p}$ be the hidden state at token position $p$, $\mu_\ell$ the average state at layer $\ell$, and $\sigma_{\ell,j}$ the spread of coordinate $j$ around that average. With $x=h_{\ell,p}-\mu_\ell$, we obtain
\begin{equation}
v=\mu_R+s_R\frac{Dx}{\lVert Dx\rVert_2},
\qquad D=\operatorname{diag}\!\left(\frac{\sigma_R}{\sigma_\ell}\right).
\label{eq:transport}
\end{equation}
Here $s_R$ is the median distance of insertion-layer states from $\mu_R$. We estimate each squared spread from the mean squared deviation and floor it at 1\% of its layer's median before taking the square root. This avoids amplifying nearly constant coordinates. All statistics come from plain-text passages unrelated to the test items, so the transport requires no training or fitting on pictures. In \suppref{supp:implementation}, we provide the passage splits and estimators.
Centering yields a token's deviation from its layer's average. The diagonal rescaling gives each coordinate its spread at the insertion layer, and normalization puts the result at the usual distance from $\mu_R$. At $\ell=R$, Equation (\ref{eq:transport}) only rescales the centered state.

\paragraph{Visualization.}
\label{sec:drawing}
A word's carried state replaces \emph{thing} in \emph{a photo of a thing}, or \emph{thing} alone for \sensenova{}, whose system prompt already asks for an image.
For a word split across tokens, we extract its final token's state. Each model then draws with its usual sampler and guidance. %
\suppref{supp:two} extends this to two hidden states in one request.

\paragraph{Lens variants.}
\label{sec:variants}
We chose the \vlens{} procedure among simpler and heavier ways of transporting a hidden state (\cref{tab:variants}). The simplest variants copy the state unchanged or apply the model's final normalization, including its gain, following Diffusion Lens~\citep{toker2024diffusion}. The heavier variants fit an average Jacobian map from each layer to $R$ with gradients on 64 prompts, once with an added gain per layer. Writing the read word (e.g., \emph{animal}) into the request without its hidden state gives a context-free reference. Each lens variant is scored by how often it visualizes the concept that a question describes without explicitly naming it, with items and scoring detailed in \suppref{sec:supp-variants}. The coordinate scaling of \cref{eq:transport} performs best on the validation set and remains best on the test set of fresh items, while requiring only simple layer statistics. Implementation details are in \suppref{supp:implementation}.

\paragraph{Lens tuning.}
\label{sec:lens-tuning}
The \vlens{} passes gradients, so a picture can supervise the hidden state it visualizes. For a target picture with representation $z$, Gaussian noise $\epsilon$, and $t$ uniform on $[0.4,1]$, we mix them as $z_t=(1-t)z+t\epsilon$, where larger $t$ means more noise. The frozen denoiser $f$ receives the drawing request $c(h_\theta)$, in which the chosen token's state $h_\theta$, computed with text adapters $\theta$ and carried by the \vlens{}, replaces \emph{thing}. \emph{Lens tuning} minimizes
\begin{equation}
\mathcal{L}_{\mathrm{picture}}
=\mathbb{E}_{z,\epsilon,t}\bigl\|f(z_t,t,c(h_\theta))-(\epsilon-z)\bigr\|_2^2.
\label{eq:supervision}
\end{equation}
We disable guidance and train only text adapters before layer $R$, which act only on the source text.

\begin{table}[t]
    \centering
    \small
    \setlength{\tabcolsep}{3pt}
    \caption{Lens variants on validation (Val) and test data. Scores give the share of correctly answered questions whose unnamed object is visualized, averaged over three models~\citep{deng2025bagel,xie2025showo2,diao2026sensenova}.}
    \label{tab:variants}
    \begin{tabular*}{\columnwidth}{@{\extracolsep{\fill}}lcccrr@{}}
        \toprule
        Variant & State & Grad.-free & Map-free & Val & Test \\
        \midrule
        Word (no state) & $\times$ & \checkmark & \checkmark & 0.16 & 0.03 \\
        Final norm~\citep{toker2024diffusion} & \checkmark & \checkmark & \checkmark & 0.17 & 0.11 \\
        Unchanged copy & \checkmark & \checkmark & \checkmark & 0.48 & 0.25 \\
        Jacobian map & \checkmark & $\times$ & $\times$ & 0.49 & 0.27 \\
        Jacobian + gain & \checkmark & $\times$ & $\times$ & 0.51 & 0.25 \\
        \vlens{}\,\vlensicon{} (ours) & \checkmark & \checkmark & \checkmark & \textbf{0.54} & \textbf{0.33} \\
        \bottomrule
    \end{tabular*}
\end{table}

\section{Experiments}
\label{sec:experiments}

We first test whether a token's visualization aligns with its context and how this alignment changes across layers. These tests use \bagel{}~\citep{deng2025bagel}, \showo{}~\citep{xie2025showo2}, \sensenova{}~\citep{diao2026sensenova}, and \hunyuan{}~\citep{hunyuan2025image3}, with a fixed insertion layer per model, and the depth study also includes the autoregressive \liquid{}~\citep{wu2024liquid}. We then examine purely autoregressive models and apply lens tuning to improve hidden-state visualizations as well as answers about object states.

\subsection{Setup}
\label{sec:setup}

We visualize each hidden state with three seeds. Three multimodal judges, Qwen3-VL~\citep{bai2025qwen3vl}, Gemma-3~\citep{gemmateam2025gemma3}, and LLaVA-OneVision-2~\citep{an2026llavaov2}, identify a picture's main subject, such as the elephant from \emph{animal} in \cref{fig:teaser}. They are asked to ignore incidental background objects and written words. An image is successful when at least two judges identify its correct subject. A state has a correct visualization if at least two of three seeds generate a successful image. On a separate labeled calibration set, the model-based judgment is correct on 146/156 pictures, including 67/72 with background distractors. Three human raters each agreed with the judgment on at least 88\% of 600 sampled pictures. \suppref{supp:judging} shows the questions, calibration, and rating breakdown.

Two tests analyze a fixed token's hidden state as its context in the text prompt changes. The first uses 30 pairs of factual questions describing unnamed objects, such as \emph{an animal with a long trunk} and \emph{a tall plant with a thick trunk}. The \vlens{} reads the last word of each description, here \emph{trunk}, and the judges check whether its picture shows the described object, an elephant or a tree. The second test uses 20 homonyms (words with at least two meanings), each appearing twice in a sentence after clues to its meaning, such as \emph{a whiskered seal} and \emph{a red wax seal}. Reversing the order of their occurrence yields 40 sentences and 80 states to visualize. \suppref{supp:items} explains how the items were built.

For descriptions, a pair is considered a success only when both texts' hidden states succeed at the same layer. For words with two meanings, both occurrences must succeed in the sentence, and the word's score averages the two orders. Sharing the final word across descriptions and reversing the order of the two meanings test whether pictures follow context, while controlling for a word's usual associations and its position in the text. For these tests, the extraction layer is the insertion layer $R$. To estimate what pictures that ignore context would score, we randomly swap the intended and alternative labels and rescore the entire test. The average over 10,000 such relabelings estimates chance performance, and the one-sided $p$-value is the share of relabelings that score at least as high as the true labels. \suppref{supp:context} also reports results at other extraction layers.

\begin{table}[t]
    \centering
    \small
    \setlength{\tabcolsep}{3pt}
    \caption{Both texts must show their intended object at the insertion layer for a pair to succeed. Scores for words with two meanings average the two sentence orders of each word.}
    \label{tab:context}
    \begin{tabular*}{\columnwidth}{@{\extracolsep{\fill}}lrrr@{}}
        \toprule
        Model & Success & Chance & $p$ \\
        \midrule
        \multicolumn{4}{@{}l}{Unnamed objects (30 description pairs)} \\
        \bagel{} & 2/30 & 0.03 & 0.25 \\
        \showo{} & 8/30 & 0.13 & 0.004 \\
        \sensenova{} & 4/30 & 0.07 & 0.064 \\
        \hunyuan{} & 6/30 & 0.10 & 0.016 \\
                \midrule
        \multicolumn{4}{@{}l}{Two meanings (20 words)} \\
        \bagel{} & 12.5/20 & 0.31 & 0.0001 \\
        \showo{} & 15/20 & 0.37 & 0.0001 \\
        \sensenova{} & 15/20 & 0.38 & 0.0001 \\
        \hunyuan{} & 15/20 & 0.38 & 0.0001 \\
        \bottomrule
    \end{tabular*}
\end{table}

\begin{figure*}[t]
  \centering
  \vspace{-0.25cm}
  \includegraphics[width=\textwidth]{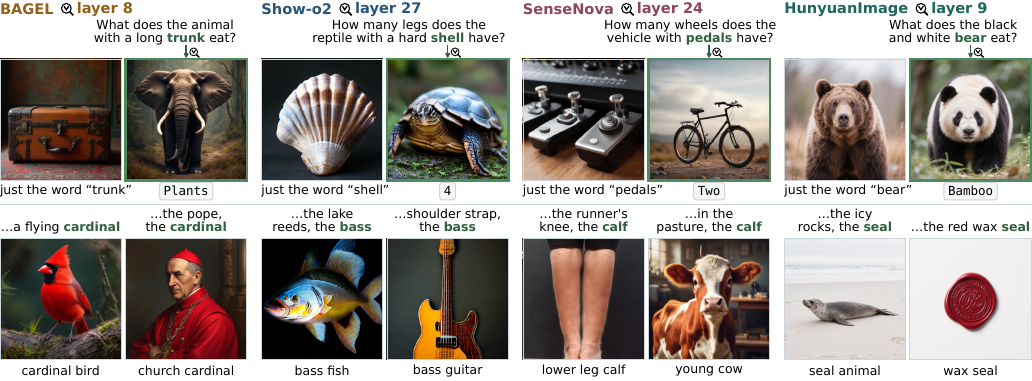} \vspace{-0.35cm}
  
  \caption{A word's hidden state can show the object its context describes, even when the word alone shows something else. Above, each pair compares the highlighted word alone with its hidden state inside the question, visualized through the \vlens{}, with the model's text answer in typewriter type. Below, one word appears twice in a sentence with two meanings, and each occurrence is visualized separately.}
  \label{fig:context}
  \vspace{-0.25cm}
\end{figure*}

\subsection{The same word in different contexts}
\label{sec:context-results}

Context can change the object visualized from the same word's hidden state (\cref{tab:context}). Words with two meanings give the most consistent result, with 12.5/20 for \bagel{} and 15/20 for each other model, all far above label-swap chance. These words already name the pictured objects, suggesting that choosing a meaning is easier to visualize than inferring an object. For unnamed objects, \showo{} recovers both subjects in 8/30 description pairs and \hunyuan{} in 6/30.
\Cref{fig:context} compares the \vlens{} reading of a word in context with the model's default drawing of that word alone.

\subsection{Intermediate concepts across tokens and layers}
\label{sec:depth}

\begin{figure*}[t]
  \centering
  \vspace{-0.2cm}\includegraphics[width=\textwidth]{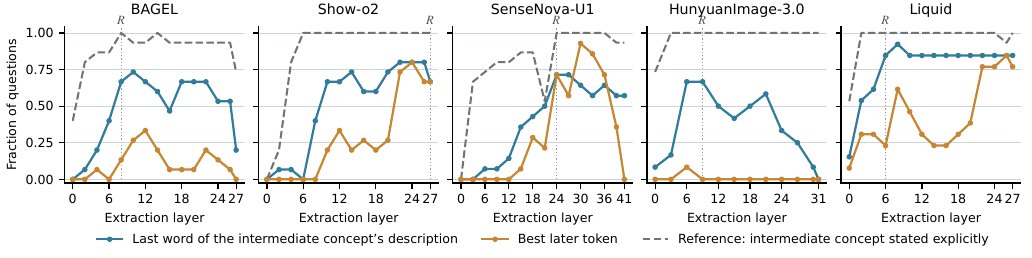}
  \caption{The object connecting a question to its answer becomes visualizable at different tokens and depths. Curves show the share of questions whose \vlens{} readings recover that object in two of three seeds. The best later token is the best token after the description's last word through the first answer token, including the instruction and chat-format tokens. The dotted lines mark each model's insertion layer.}
  \label{fig:layers}
\end{figure*}

Having established that context can change a token's \vlens{} visualization, we ask at what point in a model's computation an intermediate concept becomes visualizable. To answer this, we create sixteen two-hop reasoning questions that describe a concept (first hop) and then ask for a short answer that depends on the described concept (second hop), like the teaser question \emph{On which continent does the biggest land animal live?} (\cref{fig:teaser}). We retain questions that the model answered correctly without naming the intermediate concept and read the description's last word (\emph{animal} in the teaser) and every later token, including the instruction and chat-format tokens. As a reference, we visualize the concept's token in a plain sentence, such as \emph{Yesterday I saw an elephant on my way home.} \Cref{fig:layers} shows the shares of questions with successful visualizations at the description's last word, their best later token, and in the plain sentence. 

At the description's last word, the \vlens{} recovers the concept in at least two thirds of questions at $R$ in all five models. Later recovery is much weaker in \bagel{} and \hunyuan{}, while \showo{}, \sensenova{}, and \liquid{} often make the object visualizable at chat-format tokens. These tokens open the answer turn in \showo{} and close the empty thinking block in \sensenova{}, recovering the concept more often than any single question word, consistent with low-content positions serving as computation sites~\citep{darcet2024vision}. Even here, recovery depends on depth. Just before the answer, \showo{} falls from 11/15 at layer 24 to 0/15 at layer 27, a prespecified difference of 0.73 with a 95\% interval of $[0.53,0.93]$. The concept's token in a plain sentence remains readable in all 15 cases at both layers. Among the other models, only \liquid{} shows the concept immediately before the answer, in up to 4/13.
\suppref{supp:depth} breaks these curves down by token. Text readouts of the same hidden states (logit lens, tuned lens, J-lens) rank the object among their ten most likely words at the description's last word in at most 4 questions per model at $R$ (\suppref{supp:text-readouts}).

\subsection{Picturing a thought while copying a sentence}
\label{sec:held-thoughts}

\begin{figure}[t]
    \centering
    \includegraphics[width=\columnwidth]{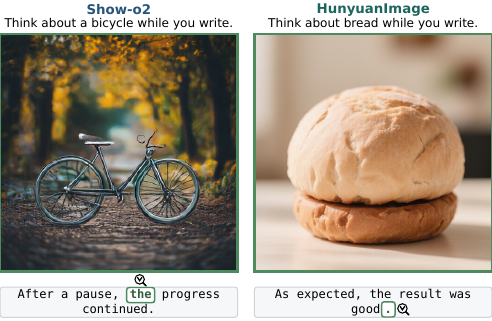}
    \caption{The model is asked to think about an object while copying an unrelated sentence. The \vlens{} can reveal this object visually from a token of the copied sentence at layer 25. Each think-instruction is shown above and the copied sentence below.}
    \label{fig:held-thoughts}
\end{figure}

Following the setup of earlier work~\citep{kowalski2026measuring,lindsey2026introspection}, we ask the two fully unified models \showo{} and \hunyuan{} to copy a sentence verbatim while thinking about an unrelated object. We then read the \vlens{} only at the tokens of the copied sentence in the response. Responses only count when their case-insensitive Ratcliff--Obershelp similarity~\citep{ratcliff1988pattern} to the original target sentence is at least 0.85.
In \showo{}, the word \emph{the} in the copy shows a bicycle at layer 25 in all three seeds when the model is asked to think about a bicycle, and in no seed when it is asked to think about a rose or receives no such instruction (\cref{fig:held-thoughts} shows one seed). In \hunyuan{}, the final period token of the copy reveals a bread roll at layer 25. These two tokens carry little content of their own, like the chat-format tokens in \cref{sec:depth}.

\subsection{Autoregressive image generators}
\label{sec:autoregressive}

\begin{figure}[t]
    \centering
    \includegraphics[width=\columnwidth]{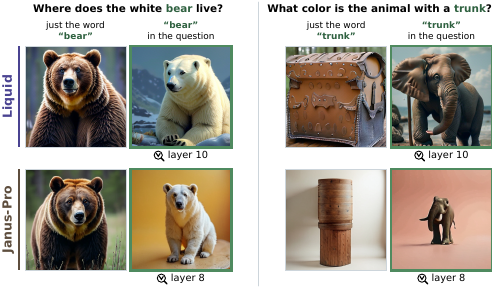}
    \caption{The \vlens{} reads contextual concepts through the purely autoregressive unified models \liquid{}~\citep{wu2024liquid} and \januspro{}~\citep{chen2025januspro}. Each pair shows the highlighted word drawn on its own (left picture of each pair) beside the visualization of its hidden state in the question through the \vlens{} (right picture of each pair).}
    \label{fig:autoregressive}
\end{figure}

The \vlens{} also reads hidden states in \liquid{}~\citep{wu2024liquid} and \januspro{}~\citep{chen2025januspro}, two unified models whose image generators produce discrete image tokens one after another. We use the same coordinate scaling and drawing request. In \cref{fig:autoregressive}, the state of \emph{bear} shows a polar bear when the question asks where the white bear lives, and that of \emph{trunk} shows the elephant even though it is not mentioned explicitly in the question. On the two-meaning test of \cref{tab:context}, both meanings are recovered for 12/20 words in \liquid{} and 11.5/20 in \januspro{}, averaging the two sentence orders. Again, both substantially exceed the chance performance ($p=0.0002$ and $p=0.0001$). \Cref{fig:layers} counts 13 questions for \liquid{}, while \januspro{} answers only 6/16 correctly (\suppref{supp:autoregressive}). %

\subsection{Supervision through the lens}
\label{sec:supervision}

We test whether the \vlens{} can visualize the referent of the pronoun \emph{it} in 20 pairs of texts that use a verb to specify which one of two named objects is referred to. For example, \emph{it} is the sandwich in \emph{A candle and a sandwich were on the table. He ate it.} and the candle when the text ends with \emph{He lit it.} At the insertion layer, the untrained models reveal both referents in 0/20 pairs for \bagel{}, 4/20 for \sensenova{}, and 0/20 for \hunyuan{}, while \showo{} correctly visualizes 16/20 (\cref{tab:supp-context}). We therefore apply lens tuning (\cref{sec:method}) to the first three models.

\begin{figure*}[t]
    \centering
    \vspace{-0.25cm}
    \includegraphics[width=\textwidth]{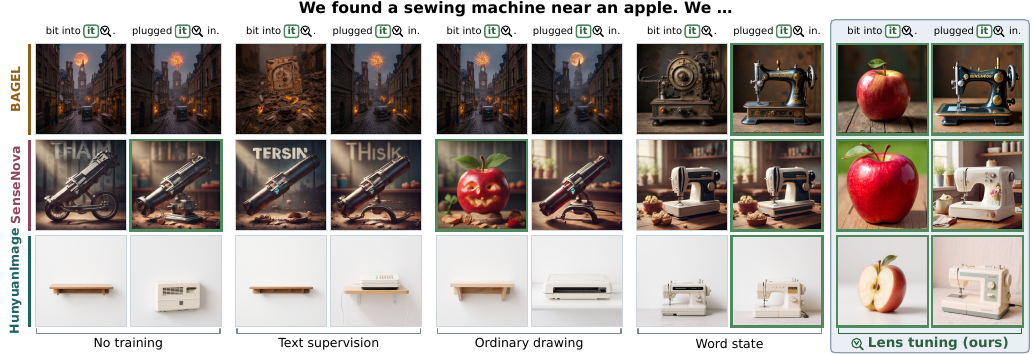}
    \caption{Lens tuning teaches a pronoun's hidden state to show the object it refers to. Changing the verb in the held-out sentence selects one of its two objects. Each column pair shows the \vlens{} pictures of the boxed \emph{it} after one kind of training, with correct pictures framed. Only lens tuning makes both pictures correct in all three models. \Cref{tab:supervision} gives scores for all test pairs and seeds.}
    \vspace{-0.25cm}
    \label{fig:supervision}
\end{figure*}

\paragraph{Training setup.}
\label{sec:supervision-setup}

Each training text also names two things and uses a verb to identify one of them. We use 48 nouns for training and 18 each for development and testing. Each model is trained on the texts in which it correctly names the referent. We evaluate on three test sets of 80 texts each, all unseen during training. The judges pick one of four options for each picture, namely the specified referent, the other named object, both, or neither. A pair succeeds only when both texts show their specified referent.
We train rank-8 low-rank adapters (LoRA) in the attention and feed-forward weights~\citep{hu2021lora}, in layers 0 to 7 for \bagel{}, 0 to 23 for \sensenova{}, and 0 to 8 for \hunyuan{}. As image targets, we use pictures the frozen model draws when prompted with the referent. Details are in \suppref{supp:training,supp:items}.

We compare lens tuning with text supervision of the referent's name, word state supervision that aligns centered pronoun and referent hidden states by cosine similarity, and ordinary drawing supervision using the same pictures and loss through the usual drawing route with \emph{A photo of it.} appended to the text. The comparisons share texts, adapter locations, and update budgets. In \cref{tab:supervision}, lens success requires both pronoun states of a pair to show their referents through the \vlens{} in two of three seeds. We evaluate lens success on unseen texts to test whether lens tuning generalizes and makes states more visually interpretable. The column \emph{Verbs} uses unseen verbs, \emph{Nouns} tests generalization on unseen nouns, and \emph{Phrases} uses sentence forms that were not encountered during training, such as \emph{Our neighbour kept a goat and a purse. One day he fed it.} %

\paragraph{What the training changes.}
\label{sec:supervision-results}

Lens tuning makes pronoun referents visualizable across the test sets (\cref{tab:supervision,fig:supervision}). Across 120 test pairs and three training seeds, success rises from 0.02 to 0.86 in \bagel{}, 0.11 to 0.94 in \sensenova{}, and 0.06 to 0.86 in \hunyuan{}. %
\begin{table}[t]
    \centering
    \footnotesize
    \setlength{\tabcolsep}{1.5pt}
    \caption{Lens success on the three test sets, averaged over three training seeds. A held-out pair succeeds when the pronoun hidden states of both its texts show their referents through the \vlens{}.}
    \resizebox{\columnwidth}{!}{%
    \begin{tabular}{@{}lccccccccc@{}}
        \toprule
        & \multicolumn{3}{c}{\bagel{}} & \multicolumn{3}{c}{\sensenova{}} & \multicolumn{3}{c}{\hunyuan{}} \\
        \cmidrule(lr){2-4}\cmidrule(lr){5-7}\cmidrule(l){8-10}
        Supervision & \scriptsize Verbs & \scriptsize Nouns & \scriptsize Phrases & \scriptsize Verbs & \scriptsize Nouns & \scriptsize Phrases & \scriptsize Verbs & \scriptsize Nouns & \scriptsize Phrases \\
        \midrule
        None & 0.00 & 0.03 & 0.03 & 0.05 & 0.13 & 0.15 & 0.00 & 0.00 & 0.18 \\
        Text & 0.00 & 0.05 & 0.02 & 0.04 & 0.23 & 0.19 & 0.00 & 0.03 & 0.17 \\
        Ordinary drawing & 0.01 & 0.13 & 0.05 & 0.25 & 0.58 & 0.51 & 0.00 & 0.01 & 0.24 \\
        Word state & 0.52 & 0.62 & 0.90 & 0.60 & 0.96 & \textbf{1.00} & 0.65 & \textbf{0.97} & \textbf{0.99} \\
        Lens tuning (ours) & \textbf{0.70} & \textbf{0.93} & \textbf{0.96} & \textbf{0.84} & \textbf{0.99} & 0.99 & \textbf{0.71} & 0.94 & 0.92 \\
        \bottomrule
    \end{tabular}}
    \vspace{-0.3cm}
    \label{tab:supervision}
\end{table}
Lens tuning succeeds on 0.92--0.99 of pairs with unseen nouns or sentence forms, while the specification of the referent through verbs remains harder with only 0.70--0.84 correctly visualized pronoun pairs (\cref{tab:supervision}). Picture supervision through the ordinary drawing request also helps visualize the hidden state, most clearly in \sensenova{}, but falls well short of lens tuning. Word state supervision also makes referents visualizable without pictures and exceeds lens tuning on unseen phrases in \sensenova{} and on unseen nouns and phrases in \hunyuan{}, but it trails lens tuning on unseen verbs in all three models and on nouns in \bagel{} and \sensenova{}. Finally, text supervision leaves lens success near its untrained level. Lens tuning shows no significant loss in general image generation on GenEval~\citep{ghosh2023geneval} or in text knowledge on MMLU-Pro~\citep{wang2024mmlupro} (\suppref{supp:capabilities}). Making a state visualizable gives us a way to supervise a unified model's reasoning through pictures. %
This motivates an investigation into whether visual supervision through the \vlens{} can also affect a model's verbal answers. 

\subsection{Training the pictured state changes the answer}
\label{sec:state-training}

\begin{figure}[t]
    \centering
    \vspace{-0.25cm}\includegraphics[width=\columnwidth]{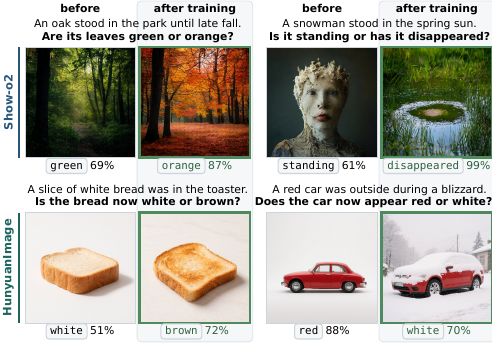}
    \caption{Training the pictured state changes the answer in \showo{} and \hunyuan{}. The \vlens{} visualizes the subject of supervised stories before and after lens tuning. Below each picture is the model's text answer with its corresponding probability.}
    \label{fig:states}
    \vspace{-0.35cm}
\end{figure}

We next ask whether lens tuning, which uses purely visual supervision, can also change a model's textual response behavior. We test the fully unified \showo{} and \hunyuan{} on 40 paired scenarios (80 stories). For example, \emph{A slice of white bread was \storydiff{put in the toaster} for three minutes.} is paired with \emph{A slice of white bread was \storydiff{left in its bag} for three minutes.} We ask \emph{Is the slice of bread now white or brown?} We supervise the second \emph{bread}, in the question, through the \vlens{} with the loss of \cref{eq:supervision}, again using the model's own pictures of brown toast for the toaster story and white bread for the bag story. Because models favor one option position, the experiment also asks each question with the options reversed, as in \emph{Is the slice of bread now brown or white?} A story counts as correct only when the one-word replies name the right option in both orders (\cref{tab:states}), and \cref{fig:states} reports a two-option probability from the margin averaged over both orders.

Training the model with pictures that show the correct state of the object raises the share of stories that \showo{} answers correctly from 36.3\% to 49.6\%, while its corresponding \vlens{} visualizations show the right state in 73.3\% instead of 52.5\% (\cref{tab:states}). Supervision with pictures that show the incorrect state leaves the answers near the untrained level and reduces the correct-picture rate to 27.1\%. In \hunyuan{}, correct answers rise from 40.0\% to 50.8\%, but training on the incorrect state also raises them to 47.5\%. \Cref{fig:states} shows the pictured state and the preferred answer changing together, as when a snowman in the spring sun disappears or white bread put in the toaster turns brown.

To further test the impact of lens tuning on the hidden states, we use a second set of 56 stories and perform isolated swaps of hidden states. Specifically, in each story we replace the subject token's state at every layer of the original \showo{} with the corresponding state from the picture-trained counterpart. We then read the same question and check whether the original model can now answer it correctly using the subject's swapped hidden state. Indeed, correct answers rise from 41/56 to 47--49/56, transferring 21--39\% of the margin gain. The trained subject states thus carry part of the answer change, though this transfer does not isolate the information shown in their pictures. Settings and per-seed results are listed in \suppref{supp:training}.

\begin{table}[t]
    \vspace{-0.25cm}
    \centering
    \small
    \setlength{\tabcolsep}{3pt}
    \caption{Lens tuning with different targets on 80 training stories. Columns are the shares (\%) of correct text answers and correct subject pictures from the \vlens{}. Means $\pm$ s.d. over three seeds.}
    \vspace{-0.1cm}
    \label{tab:states}
    \begin{tabular*}{\columnwidth}{@{\extracolsep{\fill}}llrr@{}}
        \toprule
        Model & Supervision & Correct text & Correct picture \\
        \midrule
        \multirow{3}{*}{\showo{}} & Untrained & 36.3 & 52.5 \\
        & Lens tuning (wrong) & 34.6\,$\pm$\,0.7 & 27.1\,$\pm$\,1.9 \\
        & Lens tuning (right) & \textbf{49.6}\,$\pm$\,9.5 & \textbf{73.3}\,$\pm$\,1.9 \\
        \midrule
        \multirow{3}{*}{\shortstack[l]{Hunyuan-\\Image-3.0}} & Untrained & 40.0 & 37.5 \\
        & Lens tuning (wrong) & 47.5\,$\pm$\,4.3 & 45.8\,$\pm$\,3.1 \\
        & Lens tuning (right) & \textbf{50.8}\,$\pm$\,5.6 & \textbf{64.6}\,$\pm$\,3.8 \\
        \bottomrule
    \end{tabular*}
\end{table}

\subsection{Which parts of the \vlens{} matter}
\label{sec:ablation}

To see how each part helps the \vlens{}, we repeat the two-hop test of \cref{tab:variants} on its three models, changing one part at a time. Centering removes the source layer's average, adding the insertion-layer mean sets the new center, and the two layers' spreads adjust each coordinate. Every change lowers mean recovery (\cref{tab:ablation}), least when the coordinate scaling is dropped. Removing either mean or using only the insertion spreads leaves \bagel{} and \sensenova{} near zero, while \mbox{\showo{}} retains more correct pictures.

\begin{table}[t]
    \centering
    \footnotesize
    \setlength{\tabcolsep}{1.5pt}
    \caption{Ablations of \cref{eq:transport}. Each column changes one part: using no coordinate scaling ($D{=}I$), no centering on the source mean $\mu_\ell$, no insertion-layer mean $\mu_R$, or spreads from only one layer. We count the visualized unnamed concepts as in \cref{tab:variants}.}
    \vspace{-0.1cm}
    \label{tab:ablation}
    \begin{tabular*}{\columnwidth}{@{\extracolsep{\fill}}lrrrrrr@{}}
        \toprule
        Model & \vlens{} & $D{=}I$ & No $\mu_\ell$ & No $\mu_R$ & Only $\sigma_\ell$ & Only $\sigma_R$ \\
        \midrule
        \bagel{} & \textbf{10/35} & 9/35 & 0/35 & 0/35 & 8/35 & 0/35 \\
        \showo{} & \textbf{15/30} & 13/30 & 12/30 & 11/30 & 11/30 & 9/30 \\
        \sensenova{} & \textbf{8/39} & 7/39 & 1/39 & 1/39 & 4/39 & 1/39 \\
        \midrule
        Mean & \textbf{0.33} & 0.29 & 0.14 & 0.13 & 0.23 & 0.11 \\
        \bottomrule
    \end{tabular*}
    \vspace{-0.2cm}
\end{table}

\section{Conclusion}
\label{sec:conclusion}
\vspace{-0.05cm}
The \vlens{} enables the visualization of intermediate hidden states in four unified models with diffusion-based image generators and two purely autoregressive models. Lens tuning strengthens state-referent associations, making pronoun referents visualizable in \bagel{}, \sensenova{}, and \hunyuan{}, and correcting text answers on supervised stories in \showo{} and \hunyuan{}. Transferring the trained subject word's states to the original \showo{} carries part of the answer correction, which connects the states we visualize to the answers the model writes.
What we can visualize depends on the token, layer, and drawing request. Importantly, a failure to visualize a concept does not establish that the hidden state lacks information about it. The answer corrections were measured on supervised stories, and the state transfers test whole hidden states without isolating the information visible in their pictures. The lens thus connects the inspection of intermediate hidden states with training through the model's own generator.

\section*{Acknowledgements}
The research was partially funded by a LOEWE-Spitzen-Professur (LOEWE/4a//519/05.00.002-(0010)/93). Additionally, the research was partially funded by an Alexander von Humboldt Professorship in Multimodal Reliable AI, sponsored by the Federal Ministry of Research, Technology, and Space (BMFTR). For compute, we gratefully acknowledge support from the hessian.AI Service Center (funded by the Federal Ministry of Research, Technology and Space (BMFTR), grant no. 16IS22091) and the hessian.AI Innovation Lab (funded by the Hessian Ministry for Digital Strategy and Innovation, grant no. S-DIW04/0013/003). Tobias Braun acknowledges support from ELSA (European Lighthouse on Secure and Safe AI), funded by the European Union under grant agreement No.~101070617. The views expressed are those of the authors and do not necessarily represent those of the European Union or the granting authority. Neither can be held responsible for them. Lorenzo Cavallaro was partially supported by a Google Academics Research Award.

{
    \small
    \bibliographystyle{ieeenat_fullname}
    \bibliography{references}
}
\ifwithsupp
\clearpage
\onecolumn
\appendix
\section*{Supplementary Material}
\suppressfloats[t] %
\setcounter{section}{0}
\setcounter{figure}{0}
\setcounter{table}{0}
\setcounter{equation}{0}
\renewcommand{\thesection}{\Alph{section}}
\renewcommand{\thesubsection}{\thesection.\arabic{subsection}}
\renewcommand{\thefigure}{S\arabic{figure}}
\renewcommand{\thetable}{S\arabic{table}}
\renewcommand{\theequation}{S\arabic{equation}}

\newcommand{\suppsmall}{\small\setlength{\tabcolsep}{4pt}\renewcommand{\arraystretch}{1.12}}
\newcolumntype{P}[1]{>{\raggedright\arraybackslash}p{#1}}
\makeatletter\setlength{\@fptop}{0pt}\makeatother

\noindent This supplement provides the settings, scoring rules, and detailed counts supporting the paper. \Cref{supp:implementation,supp:judging} describe how a hidden state becomes a picture and how pictures are judged. \Crefrange{supp:context}{supp:training} report detailed results of the context, depth, two-state, and training studies. \Cref{supp:items} describes how the test items were built, with examples, and \Cref{supp:galleries} provides additional pictures. \Cref{supp:autoregressive} covers the settings, two-meaning results, and depth study for autoregressive image generators, and \cref{supp:held} extends the held thoughts while copying. \Cref{supp:reproduce} explains reproduction, and \cref{supp:limitations} describes the limitations.

\section{Drawing settings and insertion layers}
\label{supp:implementation}

\subsection{Hidden states and drawing requests}
We take the hidden state entering a transformer block, before its normalization and attention, with layers numbered from zero. The extraction layer $\ell$ is where we take the state, and the insertion layer $R$ is where we place it in the drawing request. For a word split into several tokens, we take its last token. The model processes the source text separately, without drawing. The chosen hidden state then replaces the state of one token in a separate drawing request, and the layers after the insertion point recompute the request from that replacement. The single-state request is \emph{a photo of a thing} for \bagel{}, \showo{}, and \hunyuan{} and the bare word \emph{thing} for \sensenova{}. The two-state requests are \emph{a photo of a plain thing} and \emph{plain thing}, where the modifier state replaces \emph{plain} and the object or pronoun state replaces \emph{thing}. The alternative two-state template is \emph{a photo of a thing that is plain}. We use each model's drawing settings (\cref{tab:supp-models}), message markers, and system text. \bagel{} and \sensenova{} read the request into keys and values used by separate generation weights. \showo{} processes the text again at every denoising step, so the replacement is repeated in the conditional branch only.

\begin{table}[t]
\centering\suppsmall
\caption{The four models and their drawing settings. The checkpoints are BAGEL-7B-MoT, Show-o2-7B, SenseNova-U1-8B-MoT, and HunyuanImage-3.0-Instruct. $R$ is the insertion layer, and the fixed extraction layers are those of the context test (\cref{supp:context}). The Steps column gives the number of scheduled time points with time shift 3, and Seconds gives the approximate recorded time to draw one picture, on a single GPU for the first three models and on four GPUs for the much larger \hunyuan{}.}
\label{tab:supp-models}
\begin{tabular*}{\linewidth}{@{\extracolsep{\fill}}lrrrlrrrr@{}}
\toprule
Model & Text layers & Hidden size & $R$ & Fixed extraction layers & Pixels & Steps & Guidance & Seconds \\
\midrule
\bagel{} & 28 & 3584 & 8 & 6, 8, 14 & $512^2$ & 28 & 4.0 & 9 \\
\showo{} & 28 & 3584 & 27 & 20, 24, 27 & $432^2$ & 50 & 7.5 & 12.6 \\
\sensenova{} & 42 & 4096 & 24 & 20, 24, 32 & $512^2$ & 50 & 4.0 & 9.4 \\
\hunyuan{} & 32 & 4096 & 9 & 9, 12, 15 & $1024^2$ & 50 & 2.5 & 110 \\
\bottomrule
\end{tabular*}
\end{table}

\subsection{Carrying a hidden state between layers}
With $x=h_{\ell,p}-\mu_\ell$, the transported hidden state is
\begin{equation}
v=\mu_R+s_R\frac{D_{\ell,R}x}{\|D_{\ell,R}x\|_2},
\qquad D_{\ell,R}=\operatorname{diag}(\sigma_R/\sigma_\ell).
\end{equation}
All statistics come from plain-text passages unrelated to the test items. The means $\mu_\ell$ are estimated from 64 passages. The coordinate spreads $\sigma_\ell$ use those passages and 40 additional ones. For each coordinate we average $(h_{\ell,p,j}-\mu_{\ell,j})^2$, floor this value at 1\% of the layer's median coordinate variance, and take the square root. The radius $s_R$ is the median distance from $\mu_R$ over 20 separate passages. We use positions from the fifth token to the token before the final marker. At $\ell=R$ the diagonal matrix is the identity, but the centered state is still placed at the usual radius. The statistics remain fixed for all image tasks and adapter training, so the lens is never fitted on pictures.

\subsection{Comparing lens variants}
\label{sec:supp-variants}
\Cref{tab:supp-variants} compares the variants for each model. Each variant that uses a hidden state carries it to the insertion layer $R$, and all are visualized with the same seeds and requests. The row \emph{Word (no state)} places the word being read, such as \emph{animal}, into the request without its hidden state, as a context-free reference. The unchanged copy inserts $h_{\ell,p}$ at $R$ as it is. Two further variants change only the radius of the centered state $x$. Relative size inserts $\mu_R+(s_R/s_\ell)\,x$, where $s_\ell$ is the median distance from $\mu_\ell$, and percentile size gives $x$ the insertion-layer distance at the percentile that $\|x\|$ has among layer-$\ell$ distances.

The validation set has 70 two-hop questions and 45 words with two meanings, and the test set has 72 questions and 40 words, all written for this comparison. For each model, we retain the questions it answers correctly in text, leaving 44, 44, and 52 validation and 35, 30, and 39 test questions for \bagel{}, \showo{}, and \sensenova{}. Each word is read in both sentence orders, giving 90 and 80 sentences. We read the description's last word and the question's last word at seven extraction layers in \bagel{} (2, 4, 6, 8, 10, 14, 20) and \sensenova{} (8, 12, 16, 20, 24, 28, 36) and at six in \showo{} (6, 10, 14, 20, 24, 27), with three drawing seeds. The judges choose among the intended object, the word's own or other meaning, three other objects, and none, and a reading succeeds when most judges pick the intended object in two of three seeds. A two-hop question counts when any of its readings shows its connecting object, and a sentence counts when each occurrence of the word shows its own meaning at some layer.

We selected the variant on the validation set (Val) using a rule written before drawing, which chooses the simplest variant within one bootstrap standard error of the best. On the validation set the coordinate scaling exceeds every simpler variant by more than one standard error, so we did not run relative or percentile size on the test set. Coordinate scaling remains best on fresh test items (Test).

\begin{table*}[t]
    \centering
    \small
    \caption{Lens variants for each model. The two-hop columns count questions whose connecting object is visualized, and the both-meanings columns count sentences in which each meaning is visualized in context. The score is the mean of the six rates, and \cref{tab:variants} averages the three two-hop rates of each row in this table and in \cref{tab:supp-final-norm}.}
    \label{tab:supp-variants}
    \begin{tabular*}{\textwidth}{@{\extracolsep{\fill}}lrrrrrrr@{}}
        \toprule
        & \multicolumn{2}{c}{\bagel{}} & \multicolumn{2}{c}{\showo{}} & \multicolumn{2}{c}{\sensenova{}} & \\
        \cmidrule(lr){2-3}\cmidrule(lr){4-5}\cmidrule(lr){6-7}
        Variant & Two-hop & Both meanings & Two-hop & Both meanings & Two-hop & Both meanings & Score \\
        \midrule
        \multicolumn{8}{@{}l}{Val, used for the choice} \\
        Word (no state) & 7/44 & 1/90 & 7/44 & 0/90 & 9/52 & 0/90 & 0.08 \\
        Unchanged copy & 19/44 & 35/90 & 26/44 & 31/90 & 21/52 & 36/90 & 0.43 \\
        Relative size & 20/44 & 31/90 & 28/44 & 42/90 & 20/52 & 40/90 & 0.46 \\
        Percentile size & 19/44 & 35/90 & 31/44 & 43/90 & 18/52 & 35/90 & 0.46 \\
        Jacobian map & 18/44 & 27/90 & 29/44 & 44/90 & 21/52 & 36/90 & 0.44 \\
        Jacobian + gain & 23/44 & 34/90 & 28/44 & 40/90 & 19/52 & 38/90 & 0.46 \\
        \vlens{}\,\vlensicon{} (ours) & 24/44 & 31/90 & 31/44 & 50/90 & 20/52 & 33/90 & \textbf{0.48} \\
        \midrule
        \multicolumn{8}{@{}l}{Test, fresh items} \\
        Word (no state) & 0/35 & 0/80 & 2/30 & 0/80 & 1/39 & 0/80 & 0.02 \\
        Unchanged copy & 10/35 & 48/80 & 8/30 & 52/80 & 8/39 & 54/80 & 0.45 \\
        Jacobian map & 9/35 & 39/80 & 12/30 & 58/80 & 6/39 & 48/80 & 0.44 \\
        Jacobian + gain & 7/35 & 41/80 & 13/30 & 60/80 & 5/39 & 47/80 & 0.44 \\
        \vlens{}\,\vlensicon{} (ours) & 10/35 & 42/80 & 15/30 & 59/80 & 8/39 & 44/80 & \textbf{0.47} \\
        \bottomrule
    \end{tabular*}
\end{table*}

\subsection{Final normalization}
\label{supp:final-norm}

We adapt the final normalization used by Diffusion Lens~\citep{toker2024diffusion} to each model's image-generation pathway. We apply its final RMSNorm, including the learned gain, to the extracted hidden state and insert the result $u$ at $R$. We use the generation expert's normalization in \bagel{}, the shared final normalization in \showo{}, and the generation branch's normalization in \sensenova{}. A second variant inserts $\mu_R+s_Ru/\lVert u\rVert_2$, matching the radius used by the \vlens{} while retaining the normed state's direction. Both variants use the items and scoring of \cref{tab:supp-variants}, with the same positions, layers, requests, seeds, and samplers. We omit \hunyuan{}, whose drawing head normalizes the full latent grid at once.

On the test set, final normalization achieves 0.140, or 0.183 with the radius adjustment, against 0.467 for the \vlens{} (\cref{tab:supp-final-norm}), and on the validation set 0.122 and 0.223 against 0.484. On the test set, the differences from the \vlens{} are $-0.328$ with a 95\% paired bootstrap interval of $[-0.389,-0.266]$ and $-0.284$ with $[-0.343,-0.226]$. The normed states have only 0.15 of the usual insertion-layer magnitude in \bagel{}, 0.36 to 0.51 in \showo{}, and 0.08 to 0.10 in \sensenova{}. Their cosine with the insertion-layer mean is 0.33 to 0.50 in \bagel{} and \sensenova{}. These measurements suggest mismatches in both size and direction. Matching the radius improves recovery in \showo{}, whose drawing head reads the shared final normalization, but leaves recovery low in the other two models.

\begin{table}[t]
\centering\small
\setlength{\tabcolsep}{3pt}
\caption{Final normalization on the validation (Val) and test sets. Counts measure object recovery in questions and recovery of both meanings in sentences. The score averages the six rates.}
\label{tab:supp-final-norm}
\begin{tabular*}{\linewidth}{@{\extracolsep{\fill}}lrrrrrrr@{}}
\toprule
& \multicolumn{2}{c}{\bagel{}} & \multicolumn{2}{c}{\showo{}} & \multicolumn{2}{c}{\sensenova{}} & \\
\cmidrule(lr){2-3}\cmidrule(lr){4-5}\cmidrule(lr){6-7}
Variant & Two-hop & Both meanings & Two-hop & Both meanings & Two-hop & Both meanings & Score \\
\midrule
\multicolumn{8}{@{}l}{Val} \\
The \vlens{} & 24/44 & 31/90 & 31/44 & 50/90 & 20/52 & 33/90 & 0.484 \\
Final norm & 0/44 & 0/90 & 21/44 & 21/90 & 1/52 & 0/90 & 0.122 \\
Final norm at $s_R$ & 1/44 & 0/90 & 22/44 & 46/90 & 10/52 & 10/90 & 0.223 \\
\midrule
\multicolumn{8}{@{}l}{Test} \\
The \vlens{} & 10/35 & 42/80 & 15/30 & 59/80 & 8/39 & 44/80 & 0.467 \\
Final norm & 0/35 & 0/80 & 9/30 & 41/80 & 1/39 & 0/80 & 0.140 \\
Final norm at $s_R$ & 0/35 & 0/80 & 12/30 & 51/80 & 1/39 & 3/80 & 0.183 \\
\bottomrule
\end{tabular*}
\end{table}

\subsection{Choosing the insertion layer}
We copy the hidden state of a concrete word from \emph{Yesterday I saw a \{word\} on my way home.} into \emph{thing} in \emph{a photo of a thing} at the same layer, draw with seed 0, and let CLIP ViT-L/14 rank the picture against 100 concept names. The words are spider, elephant, piano, banana, rocket, pyramid, kangaroo, octopus, penguin, camel, bee, and apple. \sensenova{} is calibrated with this request too, although its drawing requests otherwise use the bare \emph{thing} because its system prompt already asks for an image. Some words also name objects in later tests, but calibration reads each word only from this plain sentence and never from a test question. The same sentence serves as the named-object reference of the depth study (\cref{supp:depth}). The insertion layer is the latest tested layer at which the most words are ranked first (\cref{tab:supp-calibration}). In \bagel{} the count drops to zero from layer 16 on and in \sensenova{} from layer 36 on, whereas \showo{} recovers all words at every layer from 14 on. \hunyuan{} recovers all 12 at layers 3, 6, and 9 and none from layer 21 onward. The chosen layers, $R=8$ for \bagel{}, $R=27$ for \showo{}, $R=24$ for \sensenova{}, and $R=9$ for \hunyuan{}, stay fixed in every experiment.

\begin{table}[t]
\centering\suppsmall
\caption{Number of the 12 calibration words that CLIP ranks first when the word's hidden state is placed into the drawing request at the same layer. Bold marks the chosen insertion layer, the latest layer with the highest count.}
\label{tab:supp-calibration}
\begin{tabular*}{\linewidth}{@{\extracolsep{\fill}}l*{12}{r}@{}}
\toprule
\bagel{} layer & 0 & 2 & 4 & \textbf{8} & 12 & 16 & 20 & 24 & 27 & & & \\
Words right & 10 & 12 & 12 & \textbf{12} & 8 & 0 & 0 & 0 & 0 & & & \\
\midrule
\showo{} layer & 0 & 2 & 4 & 6 & 8 & 10 & 12 & 14 & 16 & 20 & 24 & \textbf{27} \\
Words right & 10 & 12 & 11 & 11 & 11 & 11 & 11 & 12 & 12 & 12 & 12 & \textbf{12} \\
\midrule
\sensenova{} layer & 0 & 4 & 8 & 12 & 16 & 20 & \textbf{24} & 28 & 32 & 36 & 40 & 41 \\
Words right & 10 & 12 & 12 & 12 & 12 & 12 & \textbf{12} & 10 & 2 & 0 & 0 & 0 \\
\midrule
\hunyuan{} layer & 0 & 3 & 6 & \textbf{9} & 12 & 15 & 18 & 21 & 24 & 27 & 30 & 31 \\
Words right & 11 & 12 & 12 & \textbf{12} & 11 & 8 & 3 & 0 & 0 & 0 & 0 & 0 \\
\bottomrule
\end{tabular*}
\end{table}

\subsection{Sampling and seeds}
Inference uses bfloat16 weights and float32 coordinate statistics on A100-80GB and L40S-48GB GPUs, with a full denoising run for each picture. \bagel{}, \showo{}, and \sensenova{} use one GPU per picture, while \hunyuan{} uses four A100-80GB or six L40S-48GB GPUs. \bagel{} applies guidance over noise levels $(0.4,1]$ with global norm rescaling, \showo{} mixes its conditional and unconditional branches as usual, and \sensenova{} applies guidance throughout its schedule. The context and two-state studies draw with seeds 0, 1, and 2 and the depth study with seeds 3, 4, and 5. Adapter training uses training seeds 0, 1, and 2. Pronoun-state evaluation uses drawing seeds 0, 1, and 2, and ordinary drawing in the pronoun study uses seeds 0 and 1. The state-question and held-thought studies use the drawing seeds given in \cref{supp:state-training,supp:held}.

\section{Judging pictures}
\label{supp:judging}

\subsection{The main-subject question}
The judges are Qwen3-VL-30B-A3B-Instruct, Gemma-3-27B-it, and LLaVA-OneVision-2-8B-Instruct. Here we describe the default main-subject question with four choices. The lens-variant comparison uses the six choices of \cref{sec:supp-variants}, and the two-state and general capability studies use the questions of \cref{supp:two-scoring,supp:capabilities}. A judge sees only the picture and its answer choices, never the source sentence, layer, model, or training condition. The question is \emph{Which of these is depicted as a main subject of the picture? Ignore incidental background objects and written words.} The choices are the intended object, the alternative object, \emph{both, as main subjects}, and \emph{neither, as a main subject}, with the first two shuffled by picture identifier, followed by \emph{Answer with the letter only.} We select the letter with the highest model probability. A picture succeeds when at least two judges choose the intended object alone. Unless a study specifies otherwise, a hidden state succeeds when two of its three drawing seeds succeed. A pair succeeds only when both texts succeed at the same layer. For words with two meanings, both occurrences must succeed within one sentence. Averaging the two sentence orders gives a word score of zero, one half, or one.

Before the experiments, we checked this question on labeled pictures with one or two main subjects, a distracting background object, or neither candidate (\cref{tab:supp-judge}). The prespecified thresholds were 90\% overall and 80\% on background cases, and the panel met both. Three people each rated 600 pictures and 20 attention checks without the sentences and answered every check correctly (\cref{tab:supp-human}). Raters 1 and 2 saw the same lens-picture sample. Rater 3 saw a separate sample of 300 context-test pictures, 240 training-study pictures, and 60 ordinary drawings. On the training pictures this rater accepted 43/48 after lens tuning and 1/48 without training, agreeing with the panel on 0.912 of them with $\kappa=0.81$. On ordinary drawings the agreement is 0.767 with $\kappa=0.55$, because the rater more often treats an added object as a main subject.

\begin{table}[t]
\suppsmall
\begin{minipage}[t]{.47\linewidth}\centering
\caption{Judge-panel accuracy on labeled calibration pictures, counted per picture before aggregating over seeds or pairs.}
\label{tab:supp-judge}
\begin{tabular*}{\linewidth}{@{\extracolsep{\fill}}lrrr@{}}
\toprule
Picture type & Correct & Total & Accuracy \\
\midrule
One main subject & 24 & 24 & 1.000 \\
Two main subjects & 31 & 36 & 0.861 \\
Background object & 67 & 72 & 0.931 \\
Neither candidate & 24 & 24 & 1.000 \\
\midrule
All & 146 & 156 & 0.936 \\
\bottomrule
\end{tabular*}
\end{minipage}\hfill
\begin{minipage}[t]{.49\linewidth}\centering
\caption{Agreement between three human raters and the judge panel on whether the intended object is shown alone. Raters 1 and 2 saw the same sample, rater 3 a separate one.}
\label{tab:supp-human}
\begin{tabular*}{\linewidth}{@{\extracolsep{\fill}}lrrrr@{}}
\toprule
Rater & Pictures & Agreement & $\kappa$ & Checks \\
\midrule
1 & 600 & 0.948 & 0.84 & 20/20 \\
2 & 600 & 0.932 & 0.80 & 20/20 \\
3 & 600 & 0.880 & 0.76 & 20/20 \\
\bottomrule
\end{tabular*}
\end{minipage}
\end{table}

\subsection{Chance levels and intervals}
Context scores are compared with 10,000 label swaps within pairs, or within words with both meanings exchanged together. Each swap recomputes the paired score, the chance estimate is the mean swapped score, and the one-sided $p$-value is $(1+n_{\geq})/(1+10{,}000)$. Context intervals use 2,000 bootstrap samples of pairs or words. The depth and two-state comparisons use paired bootstrap samples of questions or object pairs. Training intervals use 10,000 bootstrap samples of whole noun pairs, keeping both noun orders together and averaging the seeds within each text pair. The training seeds therefore do not count as independent test items in these intervals.

\section{Context test, full scores}
\label{supp:context}
We report the two context tests of \cref{sec:context-results} and the pronoun test introduced in \cref{sec:supervision}. All three keep the chosen word fixed and change its context. Descriptions end in a shared part word, with neither target named. Pronoun pairs read \emph{it} after a clause that selects one of two named things. Words with two meanings occur twice in one sentence, with each meaning determined by the words before it. \Cref{tab:supp-context} gives every primary score with its chance level and the scores at the two other fixed extraction layers. For words with two meanings the first occurrence succeeds in 34, 36, and 36 of 40 sentences for \bagel{}, \showo{}, and \sensenova{}, and the second occurrence in 29, 33, and 33.

\begin{table}[t]
\centering\suppsmall
\setlength{\tabcolsep}{3.5pt}
\caption{Context-dependent recovery for all four models and context tests. The columns Right, Rate, Chance, $p$, and the bootstrap interval refer to extraction at the insertion layer. The last two columns give the rate at the two other fixed extraction layers and the share of pairs or words that are right at any of the three layers, both of which are reported as descriptive summaries only.}
\label{tab:supp-context}
\begin{tabular*}{\linewidth}{@{\extracolsep{\fill}}llrrrrlll@{}}
\toprule
Model & Test & Right & Rate & Chance & $p$ & 95\% interval & Other fixed layers & Any layer \\
\midrule
\bagel{} & Objects & 2/30 & 0.067 & 0.033 & 0.24878 & [0.000, 0.167] & L6 0.033, L14 0.167 & 0.200 \\
\bagel{} & Pronouns & 0/20 & 0.000 & 0.000 & 1.00000 & [0.000, 0.000] & L6 0.000, L14 0.050 & 0.050 \\
\bagel{} & Meanings & 12.5/20 & 0.625 & 0.312 & 0.00010 & [0.475, 0.775] & L6 0.300, L14 0.450 & 0.650 \\
\midrule
\showo{} & Objects & 8/30 & 0.267 & 0.133 & 0.00390 & [0.133, 0.433] & L20 0.067, L24 0.267 & 0.333 \\
\showo{} & Pronouns & 16/20 & 0.800 & 0.399 & 0.00010 & [0.600, 0.950] & L20 0.700, L24 0.900 & 0.900 \\
\showo{} & Meanings & 15/20 & 0.750 & 0.374 & 0.00010 & [0.575, 0.900] & L20 0.700, L24 0.775 & 0.800 \\
\midrule
\sensenova{} & Objects & 4/30 & 0.133 & 0.067 & 0.06429 & [0.033, 0.267] & L20 0.067, L32 0.133 & 0.167 \\
\sensenova{} & Pronouns & 4/20 & 0.200 & 0.100 & 0.05739 & [0.050, 0.400] & L20 0.050, L32 0.200 & 0.350 \\
\sensenova{} & Meanings & 15/20 & 0.750 & 0.375 & 0.00010 & [0.575, 0.900] & L20 0.650, L32 0.650 & 0.775 \\
\midrule
\hunyuan{} & Objects & 6/30 & 0.200 & 0.100 & 0.01610 & [0.067, 0.367] & L12 0.200, L15 0.233 & 0.233 \\
\hunyuan{} & Pronouns & 0/20 & 0.000 & 0.000 & 1.00000 & [0.000, 0.000] & L12 0.050, L15 0.000 & 0.050 \\
\hunyuan{} & Meanings & 15/20 & 0.750 & 0.375 & 0.00010 & [0.575, 0.925] & L12 0.750, L15 0.750 & 0.825 \\
\bottomrule
\end{tabular*}
\end{table}

\paragraph{Pronoun reference in detail.}
\Cref{tab:supp-pronoun} splits the pronoun test into the ten pairs that change the selecting verb and the ten that keep it while changing a named thing, and also includes two controls. Writing the word \emph{it} itself into the drawing request, without any hidden state, never shows the intended object. Keeping only the selecting clause, as in \emph{She sheared it.}, and reading \emph{it} from that clause usually produces pictures that show neither candidate object.

\begin{table}[t]
\centering\suppsmall
\caption{Pronoun reference by subtype, with its two controls. The middle column counts judged pictures that show the intended object when the word \emph{it} is written into the request instead of a state. The right block gives the judges' picture-level decisions when the two named things are removed from the text and only the selecting clause remains as the source text.}
\label{tab:supp-pronoun}
\begin{tabular*}{\linewidth}{@{\extracolsep{\fill}}lrrrrrrr@{}}
\toprule
& \multicolumn{2}{c}{Pairs right at $R$} & Word \emph{it} alone & \multicolumn{4}{c}{Selecting clause alone, 90 pictures} \\
\cmidrule(lr){2-3}\cmidrule(lr){4-4}\cmidrule(l){5-8}
Model & Verb changed & Verb held & Intended object & Intended & Alternative & Both & Neither \\
\midrule
\bagel{} & 0/10 & 0/10 & 0/70 & 9 & 5 & 1 & 75 \\
\showo{} & 10/10 & 6/10 & 0/61 & 27 & 9 & 3 & 51 \\
\sensenova{} & 3/10 & 1/10 & 0/70 & 14 & 12 & 1 & 63 \\
\bottomrule
\end{tabular*}
\end{table}

\section{The described object across layers}
\label{supp:depth}
Sixteen questions ask for a short answer connected to a concrete object that the question describes but does not name, such as the spider in \emph{How many legs does the animal that spins webs have?} Each question is paired with a version that names the object in a preceding sentence. A separate control reads the object's name from \emph{Yesterday I saw a \{name\} on my way home.} and shows whether a plainly named word can be visualized at each layer. The 16 questions involve 15 distinct objects, as two of them ask about the bee, so this reference contains 15 objects for every model.

We first examine three fixed positions for the prespecified depth comparison. The description position is the last token of the description's last word. The question-end position is the question's last token, its question mark. For the position before the answer, the model answers the question followed by \emph{Answer with one word.} We locate the first accepted answer in the greedy reply and read the token right before it. In \bagel{} this is the start-of-turn marker, in \showo{} the line break after the assistant marker, in \sensenova{} the blank line after its empty thinking block, and in \hunyuan{} a space, except for one question whose reply begins with bold markup. A question counts when its answer is accepted and the model has not named the object before answering, so denominators differ slightly by model and condition. The prespecified comparison tests \showo{} before the answer at layer 24 against layer 27 (\cref{tab:supp-depth-pairs}), requiring recovery in at least half the questions at layer 24 and a paired 95\% interval above zero. \Cref{fig:supp-depth} shows all curves, and \cref{tab:supp-depth} gives \showo{}'s late-layer counts.

\begin{figure}[t]
\centering
\includegraphics[width=.78\linewidth]{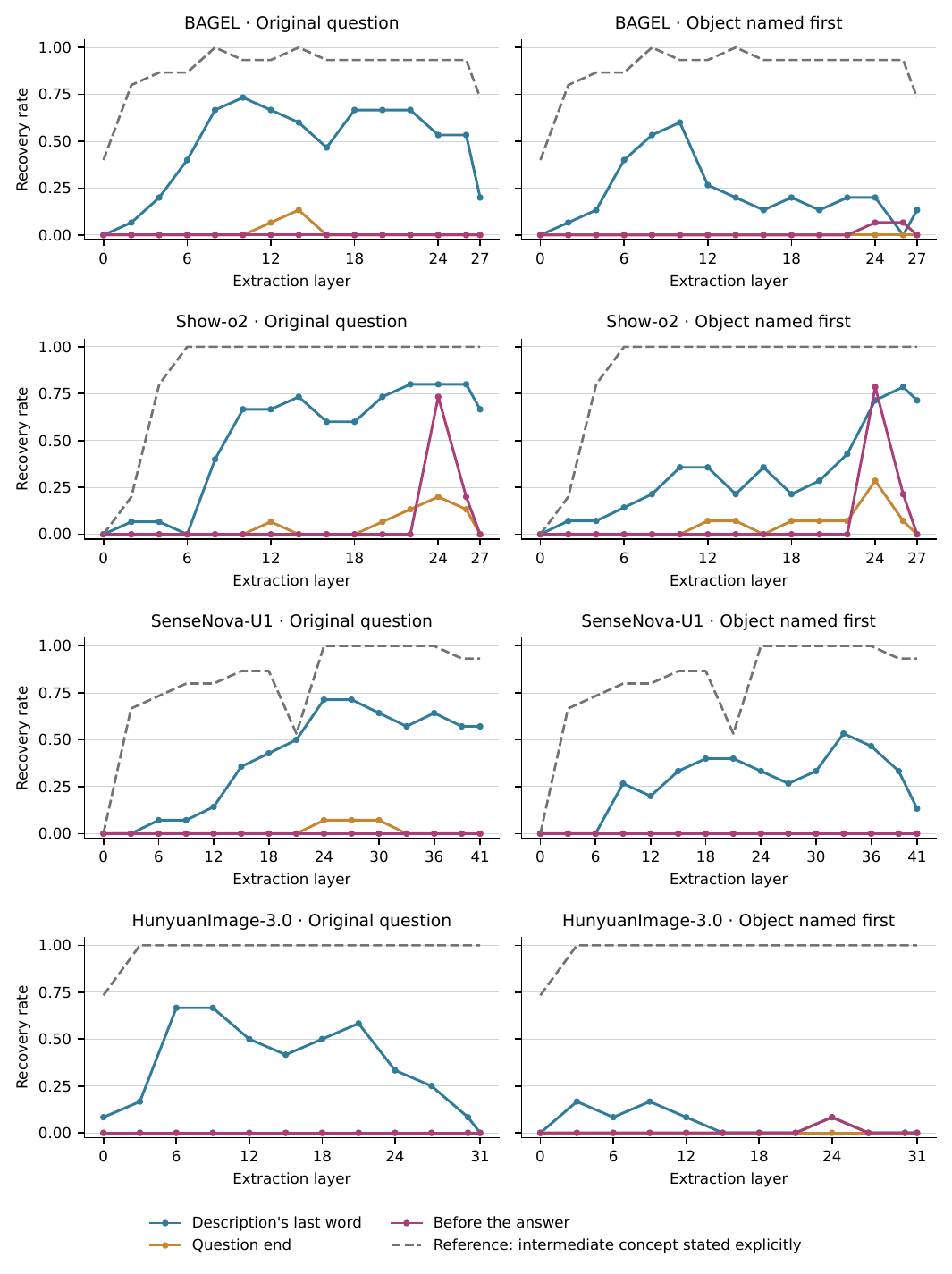}
\caption{Where the described object is readable, for the original question (left) and the version that names the object first (right). The solid curves give the share of accepted questions whose pictures show the object in two of three drawing seeds at the description's last word, the question mark, and the position before the answer. The dashed curve gives the share of the 15 distinct objects visualized from their names in plain sentences. In \showo{} the position before the answer shows the object at layer 24 and no longer shows it at layer 27, in both versions of the question.}
\label{fig:supp-depth}
\end{figure}

\begin{table}[t]
\suppsmall
\begin{minipage}[t]{.47\linewidth}\centering\setlength{\tabcolsep}{3pt}
\caption{Counts for \showo{} at its last extraction layers. The question rows count accepted questions whose picture shows the described object, and the final row counts the 15 distinct objects of the plain-sentence reference. The complete grids for all four models and both question versions are plotted in \cref{fig:supp-depth}.}
\label{tab:supp-depth}
\begin{tabular*}{\linewidth}{@{\extracolsep{\fill}}lrrrrr@{}}
\toprule
\showo{} layer & 20 & 22 & 24 & 26 & 27 \\
\midrule
Description's last word (of 15) & 11 & 12 & 12 & 12 & 10 \\
Question end (of 15) & 1 & 2 & 3 & 2 & 0 \\
Before the answer (of 15) & 0 & 0 & 11 & 3 & 0 \\
\midrule
Named first, description's last word (of 14) & 4 & 6 & 10 & 11 & 10 \\
Named first, question end (of 14) & 1 & 1 & 4 & 1 & 0 \\
Named first, before the answer (of 14) & 0 & 0 & 11 & 3 & 0 \\
\midrule
Name in a plain sentence (of 15) & 15 & 15 & 15 & 15 & 15 \\
\bottomrule
\end{tabular*}
\end{minipage}\hfill
\begin{minipage}[t]{.51\linewidth}\centering\setlength{\tabcolsep}{3pt}
\caption{Paired comparisons at the position before the answer in \showo{}, bootstrapped over questions. Comparisons across the two versions keep only questions counted in both.}
\label{tab:supp-depth-pairs}
\begin{tabular*}{\linewidth}{@{\extracolsep{\fill}}lrl@{}}
\toprule
Comparison & $n$ & Difference, 95\% interval \\
\midrule
Original, layer 24 minus 27 & 15 & +0.733 [+0.533, +0.933] \\
Named first, layer 24 minus 27 & 14 & +0.786 [+0.571, +1.000] \\
Original minus named, layer 24 & 14 & $-$0.071 [$-$0.286, +0.143] \\
Original minus named, layer 27 & 14 & +0.000 [+0.000, +0.000] \\
\bottomrule
\end{tabular*}
\end{minipage}
\end{table}

\paragraph{Every later token.}
For the best-later-token curves in \cref{fig:layers}, we also read every token after the description's last word through the first answer token. This includes the remaining question words, punctuation, the instruction \emph{Answer with one word.}, and the model's chat-format tokens. At each layer, a question counts when at least one of these hidden states shows the described object. \Cref{tab:supp-depth-tokens} separates the successful states by token type. In \showo{}, the reply-opening token \texttt{<|im\_start|>} accounts for 33 successful states, while \texttt{</think>}, which closes the empty thinking block, accounts for 59 in \sensenova{}. At layer 24, chat tokens and the line break before the answer each recover 11/15 questions in \showo{}, while chat tokens recover 13/14 at layer 30 in \sensenova{}. This closing thinking token differs from the blank line immediately before the answer tested above. In \bagel{}, most successful states come from question words such as \emph{chase} and \emph{have}, while \hunyuan{} has one successful state at \emph{have} in layer 6. Instruction tokens rarely show the object, and the first answer token usually shows neither candidate, though it can show a concrete answer such as \emph{milk}.
\begin{table}[t]
\centering\suppsmall
\setlength{\tabcolsep}{3pt}
\caption{Successful hidden states after the description, summed over questions and tested layers. A hidden state succeeds when two of three drawing seeds show the object under the three-judge rule. Several tokens can succeed for the same question and layer. The token just before the answer is counted separately from other chat-format tokens. The models contribute 15, 15, 14, and 12 questions, respectively.}
\label{tab:supp-depth-tokens}
\begin{tabular*}{\linewidth}{@{\extracolsep{\fill}}lrrrr@{}}
\toprule
Token type & \bagel{} & \showo{} & \sensenova{} & \hunyuan{} \\
\midrule
Question words & 18 & 59 & 5 & 1 \\
Question mark & 3 & 9 & 3 & 0 \\
Instruction & 0 & 2 & 0 & 0 \\
Chat format & 0 & 54 & 71 & 0 \\
Immediately before answer & 0 & 14 & 0 & 0 \\
First answer token & 8 & 5 & 1 & 0 \\
\bottomrule
\end{tabular*}
\end{table}
\subsection{Text readouts of the same states}
\label{supp:text-readouts}
We apply three text readouts to the hidden states from the context tests at $R$ and the depth study. The logit lens applies the model's final normalization and word prediction layer to the state~\citep{nostalgebraist2020logitlens}. The tuned lens first adds a learned affine map per layer~\citep{belrose2023tunedlens}, and the text J-lens first carries the state to the last layer with an average Jacobian~\citep{gurnee2026workspace}. Both are fitted on 200 plain-text passages, and \hunyuan{} has no J-lens. A reading succeeds when the highest-ranked first token across all spellings of the object's name is among the ten most likely tokens. Pairs, words with two meanings, and questions are counted as for the pictures. One description pair is left out because its two names start with the same token. \Cref{tab:supp-text-readouts} compares the counts with the \vlens{}. At the insertion layer, the text readouts rarely place the object's name among their ten most likely tokens. At the best later token, usually a chat-format token, the logit and J-lenses find it in most \sensenova{} questions, where the pictures also show it. The logit lens does so in \hunyuan{}, where the pictures do not.

\begin{table}[t]
\centering\suppsmall\setlength{\tabcolsep}{3pt}
\caption{Object recovery by the \vlens{} and by text readouts of the same states (\hunyuan{} has no J-lens). A text readout succeeds when the first token of at least one spelling of the object's name is among its ten most likely tokens. At $R$, the context tests count description pairs, pronoun pairs, and words with two meanings. The depth study counts questions at $R$ for the description's last word and at the best later token and layer.}
\label{tab:supp-text-readouts}
\begin{tabular*}{\linewidth}{@{\extracolsep{\fill}}lrrrrr@{}}
\toprule
& \multicolumn{3}{c}{Context tests at $R$} & \multicolumn{2}{c}{Depth study} \\
\cmidrule(lr){2-4}\cmidrule(l){5-6}
Readout & Descriptions & Pronouns & Meanings & At $R$ & Later \\
\midrule
\multicolumn{6}{@{}l}{\bagel{}} \\
\vlens{} & 2/30 & 0/20 & 12.5/20 & 10/15 & 5/15 \\
Logit lens & 0/29 & 0/20 & 0/20 & 0/15 & 5/15 \\
Tuned lens & 0/29 & 0/20 & 0/20 & 0/15 & 3/15 \\
J-lens & 0/29 & 0/20 & 0/20 & 1/15 & 5/15 \\
\midrule
\multicolumn{6}{@{}l}{\showo{}} \\
\vlens{} & 8/30 & 16/20 & 15/20 & 10/15 & 12/15 \\
Logit lens & 0/29 & 0/20 & 0/20 & 0/15 & 7/15 \\
Tuned lens & 0/29 & 0/20 & 0/20 & 0/15 & 5/15 \\
J-lens & 0/29 & 0/20 & 0/20 & 0/15 & 8/15 \\
\midrule
\multicolumn{6}{@{}l}{\sensenova{}} \\
\vlens{} & 4/30 & 4/20 & 15/20 & 10/14 & 13/14 \\
Logit lens & 0/29 & 0/20 & 0/20 & 0/14 & 13/14 \\
Tuned lens & 0/29 & 0/20 & 0/20 & 0/14 & 6/14 \\
J-lens & 0/29 & 0/20 & 2/20 & 4/14 & 14/14 \\
\midrule
\multicolumn{6}{@{}l}{\hunyuan{}} \\
\vlens{} & 6/30 & 0/20 & 15/20 & 8/12 & 1/12 \\
Logit lens & 0/29 & 0/20 & 1/20 & 1/12 & 8/12 \\
Tuned lens & 0/29 & 0/20 & 0/20 & 0/12 & 6/12 \\
\bottomrule
\end{tabular*}
\end{table}

\section{Reading two hidden states}
\label{supp:two}
\label{sec:appearance}

The two-state experiments use unchanged copies to visualize an object's color or pattern from two hidden states (\cref{fig:two-states}). Both states enter $R$ without the centering and rescaling of \cref{eq:transport}. Eight development pairs and eight held-out pairs with new objects and sentence forms assign two colors or patterns to each object. After \emph{The car was painted bright red. Later I looked at it.}, we copy the states of \emph{red} and \emph{it} unchanged into separate slots at the insertion layer. The modifier and noun states from an explicit drawing request check that both can be recovered together. We compare this with the pronoun alone, the pronoun in both slots, and states exchanged between texts with another color, pattern, or object.

All models recover the modifier and object from the explicit drawing requests, except for one \hunyuan{} development pair. With both required in the picture, the modifier and pronoun from contextual texts succeed in 5/8 held-out pairs for \showo{}, 3/7 for \hunyuan{}, 1/7 for \sensenova{}, and 0/7 for \bagel{}. Neither the pronoun alone nor the duplicated pronoun succeeds, and the noun alone almost never shows a value. In \hunyuan{} the noun alone already shows the colored car and cup, one pair per split. In \showo{} the gain over the pronoun alone is 0.63, with a 95\% interval of $[0.25,0.88]$. \sensenova{} often shows the intended color on a different object. Exchanging the modifier changes the color with the object present in 3/8 held-out \showo{} pairs. Exchanging the pronoun changes the object in 13/16 held-out value cases, using the main-subject question with the two objects as choices. These comparisons show that the modifier and pronoun states make separate contributions to \showo{}'s pictures. Together, the two states can show a colored object even though neither expresses both properties.

\begin{figure}[t]
    \centering
    \includegraphics[width=\linewidth]{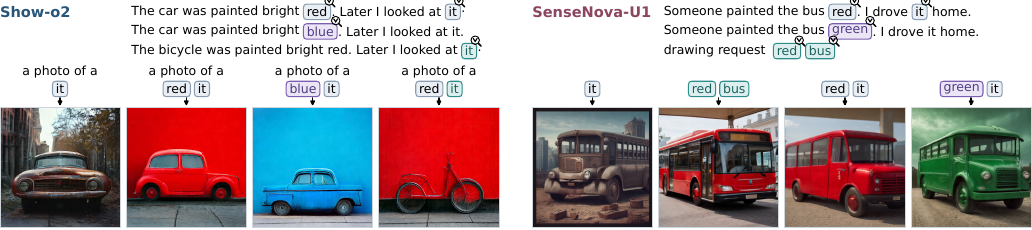}
    \caption{Combining two hidden states can make an object's color visible when the pronoun alone does not. Each request shows the states of the marked words, with colored labels marking their source sentences. In \showo{}, exchanging the color word or the pronoun changes the pictured color or object. The \sensenova{} panel includes a comparison with \emph{red} and \emph{bus} from an explicit drawing request and its one held-out pair that succeeds with the object present. Across the held-out pairs, \showo{} recovers both color and object most consistently.}
    \label{fig:two-states}
\end{figure}

\subsection{Conditions and scoring}
\label{supp:two-scoring}
We did not inspect the held-out pairs during development. Partner objects share their modifier values, so object states can be exchanged between them. The same three judges answer two questions about each picture. The attribute question is \emph{What colour is the car in this picture?}, or the pattern equivalent with \emph{stripes} and \emph{spots (polka dots)}, with the choices being the two values, \emph{both or unclear}, \emph{something else}, and \emph{there is no car}. The presence question is \emph{Is there a car in this picture?} with yes and no. A picture is correct when two of three judges give the intended value and two of three say the object is there. A state succeeds in at least two of three seeds, and a pair succeeds for both values. We include a pair when both explicit requests pass this test, which leaves 14 pairs for \bagel{} (7 development, 7 held-out), 14 for \showo{} (6 and 8), and 15 each for \sensenova{} and \hunyuan{} (8 and 7). \Cref{tab:supp-two} reports every condition scored by both values, including alternative drawing templates and two states averaged into one slot. The pronoun exchange is scored by the main-subject question instead and is reported in the text. All four models use the same items, seeds, judges, and rules.

\begin{table}[t]
\centering\suppsmall
\caption{Pairs whose pictures show both color or pattern values on the intended object, with the object required. The top block shows the controls from explicit requests and plain sentences, the middle block uses states from the contextual texts, and the bottom block tests a different template or averages two states in one slot. Conditions that use states copy them unchanged at $R$, with averaging only in the final row. When the modifier is exchanged, it comes from the same object's text with the other value, so a correct result means that the color followed the modifier state. The pronoun exchange uses the main-subject question and is reported in the text.}
\label{tab:supp-two}
\begin{tabular*}{\linewidth}{@{\extracolsep{\fill}}lrrrrrrrr@{}}
\toprule
& \multicolumn{2}{c}{\bagel{}} & \multicolumn{2}{c}{\showo{}} & \multicolumn{2}{c}{\sensenova{}} & \multicolumn{2}{c}{\makebox[0pt]{\hunyuan{}}} \\
\cmidrule(lr){2-3}\cmidrule(lr){4-5}\cmidrule(lr){6-7}\cmidrule(l){8-9}
Condition & Dev & Held-out & Dev & Held-out & Dev & Held-out & Dev & Held-out \\
\midrule
Explicit request drawn as text & 7/7 & 7/7 & 6/6 & 8/8 & 8/8 & 7/7 & 8/8 & 7/7 \\
Modifier and noun states from the explicit request & 7/7 & 7/7 & 6/6 & 8/8 & 8/8 & 7/7 & 7/8 & 7/7 \\
Modifier and noun states from a plain sentence & 7/7 & 7/7 & 6/6 & 8/8 & 8/8 & 7/7 & 8/8 & 7/7 \\
Noun state alone & 0/7 & 0/7 & 0/6 & 0/8 & 1/8 & 0/7 & 1/8 & 1/7 \\
\midrule
Modifier and pronoun states from the text & 1/7 & 0/7 & 2/6 & 5/8 & 0/8 & 1/7 & 2/8 & 3/7 \\
Pronoun state alone & 0/7 & 0/7 & 0/6 & 0/8 & 0/8 & 0/7 & 0/8 & 0/7 \\
Pronoun state in both slots & 0/7 & 0/7 & 0/6 & 0/8 & 0/8 & 0/7 & 0/8 & 0/7 \\
Modifier from the other text, same pronoun & 1/7 & 1/7 & 1/6 & 3/8 & 0/8 & 1/7 & 3/8 & 3/7 \\
\midrule
Explicit states in the alternative template & 7/7 & 7/7 & 6/6 & 8/8 & 8/8 & 7/7 & 7/8 & 7/7 \\
Text states in the alternative template & 0/7 & 0/7 & 3/6 & 4/8 & 0/8 & 0/7 & 2/8 & 3/7 \\
Average of the two explicit states in one slot & 2/7 & 0/7 & 1/6 & 0/8 & 5/8 & 5/7 & 1/8 & 0/7 \\
\bottomrule
\end{tabular*}
\end{table}

The paired gain over either the pronoun alone or the pronoun in both slots is $+0.625$ with 95\% interval $[+0.250, +0.875]$ for \showo{} and $+0.429$ $[+0.143, +0.857]$ for \hunyuan{}, so both intervals exclude zero. It is $+0.143$ $[+0.000, +0.429]$ for \sensenova{} and $0.000$ $[0.000, 0.000]$ for \bagel{}.

\section{Training through the lens}
\label{supp:training}

\subsection{Data and training}
We train adapters in \bagel{}, \sensenova{}, and \hunyuan{}. During lens tuning, adapters act only on the source text, while the drawing request and generator remain frozen. The texts use 84 noun labels in six categories, each category with eight training, three development, and three held-out labels (\cref{tab:supp-nouns}). Only categories whose selecting verbs cannot apply to the other object are paired. Each pair appears in both noun orders, with each noun selected in turn. For \bagel{} and \sensenova{}, the training pool has 60 pairs in four sentence forms, 960 texts in all. For each model, we retain the texts whose referent it names correctly, 696 for \bagel{} and 798 for \sensenova{}. Three test sets of 20 noun pairs, named as in \cref{tab:supervision}, use new pairs of training nouns with a held-out selecting verb per category (\emph{Verbs}), held-out noun labels (\emph{Nouns}), and new pairs of training nouns in two held-out sentence forms (\emph{Phrases}). Each test set has 80 texts, forming 40 pairs with the same noun order and a different selected referent. Both texts must give the intended referent. For development, we use 20 additional pairs whose noun labels do not occur in training or testing.

For \bagel{} and \sensenova{}, LoRA adapters of rank 8 and scaling 16 modify the query, key, value, attention-output, gate, up, and down projections of the understanding weights in layers 0 to 7 and 0 to 23, respectively. These models use AdamW with learning rate $10^{-4}$ and no weight decay for 600 updates of 16 texts sampled with replacement. We evaluate the final checkpoint with no selection on test data. For each training noun, we keep up to four target pictures from the frozen model's own drawings with seeds 0 to 5 when at least two judges recognize the noun. For each loss evaluation, we sample a target, a noise level uniformly on $[0.4,1]$, and fresh Gaussian noise, and use the conditional denoising loss without guidance on VAE latents for \bagel{} and pixels for \sensenova{}. Gradients pass through the frozen generator to the text adapters. During lens tuning the adapters act only on the source text, and the drawing request uses unchanged weights.

For \hunyuan{}, we place adapters in layers 0 to 8 before insertion layer 9 and train them for 200 updates of 16 texts. The four training arms use the same test items, measures, scorer, and three training seeds as the other models. Lens tuning succeeds on 0.71 of the pairs in \emph{Verbs}, 0.94 in \emph{Nouns}, and 0.92 in \emph{Phrases}. Word state supervision gives 0.65, 0.97, and 0.99. Its overall gain over text supervision is 0.79 with a 95\% interval of $[0.73,0.84]$. Ordinary drawing supervision raises pair-level drawing success from 0.03 to 0.69. Answers on the untrained question forms average 0.71 before training, 0.79 after text supervision, 0.75 after lens tuning, and 0.79 after ordinary drawing.

\subsection{Kinds of training and evaluation}
Text supervision predicts \emph{The \{referent\}.} after a question asking what \emph{it} refers to. Lens tuning applies the denoising loss through the pronoun's hidden state. Ordinary drawing supervision uses the same targets and noise schedule but conditions on the full text followed by \emph{A photo of it.}, with adapters active throughout. We use two controls. Word state supervision minimizes one minus the cosine between the centered states of the pronoun and the referent word over four neutral sentences, while the wrong-picture control applies lens tuning with pictures of a fixed wrong noun from another category.

Lens evaluation visualizes each pronoun state with three seeds. A pair of texts succeeds when each referent is shown in two of its three seeds. Ordinary drawing uses the full text followed by \emph{A photo of it.}, and a text must succeed in both seeds. Text evaluation asks the trained naming question, an untrained choice between the two nouns, and two yes or no questions about the categories of the referent and the alternative. The untrained score averages the latter three. The preservation tests cover texts that the untrained lens already visualizes correctly and 40 general questions. The prespecified transfer rule requires a gain of at least 0.15 over no training on the three test sets, an interval above zero, and an improvement on held-out noun labels. \Cref{tab:supp-train-lens,tab:supp-train-text,tab:supp-train-draw,tab:supp-train-seeds,tab:supp-train-pairs} report all \bagel{} and \sensenova{} scores by training type, test set, and seed.

\begin{table}[t]
\centering\suppsmall
\caption{Lens success after each kind of training, averaged over training seeds, on development pairs and the three test sets of \cref{tab:supervision} (held-out verbs, held-out nouns, new sentence forms). A pair succeeds if both pronoun pictures show the selected object.}
\label{tab:supp-train-lens}
\begin{tabular*}{\linewidth}{@{\extracolsep{\fill}}lrrrrrrrrrr@{}}
\toprule
& \multicolumn{5}{c}{\bagel{}} & \multicolumn{5}{c}{\sensenova{}} \\
\cmidrule(lr){2-6}\cmidrule(l){7-11}
Supervision & Dev & Verbs & Nouns & Phrases & All tests & Dev & Verbs & Nouns & Phrases & All tests \\
\midrule
None & 0.000 & 0.000 & 0.025 & 0.025 & 0.017 & 0.050 & 0.050 & 0.125 & 0.150 & 0.108 \\
Text & 0.050 & 0.000 & 0.050 & 0.017 & 0.022 & 0.075 & 0.042 & 0.233 & 0.192 & 0.156 \\
Lens tuning & 0.908 & 0.700 & 0.925 & 0.958 & 0.861 & 1.000 & 0.842 & 0.992 & 0.992 & 0.942 \\
Ordinary drawing & 0.233 & 0.008 & 0.133 & 0.050 & 0.064 & 0.575 & 0.250 & 0.575 & 0.508 & 0.444 \\
Word state & 0.642 & 0.517 & 0.617 & 0.900 & 0.678 & 0.975 & 0.600 & 0.958 & 1.000 & 0.853 \\
Wrong pictures & 0.000 & 0.000 & 0.000 & 0.000 & 0.000 & 0.000 & 0.000 & 0.000 & 0.000 & 0.000 \\
\bottomrule
\end{tabular*}
\end{table}

\begin{table}[t]
\centering\suppsmall
\caption{Text answers and preservation after each kind of training. Naming uses the trained question, while untrained averages the two-way choice and the two category questions. Retained is the share of texts that the untrained lens already visualized correctly (19 for \bagel{}, 105 for \sensenova{}) and that still succeed, and general is accuracy on 40 general factual questions.}
\label{tab:supp-train-text}
\begin{tabular*}{\linewidth}{@{\extracolsep{\fill}}lrrrrrrrr@{}}
\toprule
& \multicolumn{4}{c}{\bagel{}} & \multicolumn{4}{c}{\sensenova{}} \\
\cmidrule(lr){2-5}\cmidrule(l){6-9}
Supervision & Naming & Untrained & Retained & General & Naming & Untrained & Retained & General \\
\midrule
None & 0.779 & 0.758 & 1.000 & 0.975 & 0.829 & 0.817 & 1.000 & 0.975 \\
Text & 0.972 & 0.910 & 0.789 & 0.975 & 0.974 & 0.910 & 0.787 & 0.950 \\
Lens tuning & 0.735 & 0.764 & 0.982 & 0.975 & 0.918 & 0.776 & 0.965 & 1.000 \\
Ordinary drawing & 0.782 & 0.851 & 0.860 & 0.975 & 0.947 & 0.969 & 0.873 & 0.967 \\
Word state & 0.537 & 0.580 & 0.772 & 0.975 & 0.050 & 0.481 & 0.924 & 0.958 \\
Wrong pictures & 0.487 & 0.600 & 0.018 & 0.975 & 0.608 & 0.673 & 0.022 & 0.975 \\
\bottomrule
\end{tabular*}
\end{table}

\begin{table}[t]
\centering\suppsmall
\caption{Ordinary drawing from the full text after each kind of training, on the three test sets. A text succeeds when both drawing seeds show its referent alone, and a pair additionally requires both referents. The \emph{None} rows are the untrained models, the wrong-picture row of \bagel{} uses a single training seed, and all other rows average the three training seeds of each model.}
\label{tab:supp-train-draw}
\begin{tabular*}{\linewidth}{@{\extracolsep{\fill}}lrrrrrrrrrr@{}}
\toprule
& \multicolumn{5}{c}{\bagel{}} & \multicolumn{5}{c}{\sensenova{}} \\
\cmidrule(lr){2-6}\cmidrule(l){7-11}
Supervision & Verbs & Nouns & Phrases & All texts & Pairs & Verbs & Nouns & Phrases & All texts & Pairs \\
\midrule
None & 0.425 & 0.375 & 0.338 & 0.379 & 0.008 & 0.463 & 0.463 & 0.338 & 0.421 & 0.075 \\
Text & 0.454 & 0.450 & 0.329 & 0.411 & 0.022 & 0.446 & 0.463 & 0.254 & 0.388 & 0.047 \\
Lens tuning & 0.529 & 0.475 & 0.471 & 0.492 & 0.094 & 0.533 & 0.550 & 0.529 & 0.537 & 0.146 \\
Ordinary drawing & 0.942 & 0.912 & 0.925 & 0.926 & 0.854 & 0.975 & 0.946 & 0.992 & 0.971 & 0.942 \\
Word state & 0.504 & 0.463 & 0.487 & 0.485 & 0.064 & 0.479 & 0.550 & 0.450 & 0.493 & 0.102 \\
Wrong pictures & 0.375 & 0.375 & 0.362 & 0.371 & 0.000 & 0.433 & 0.442 & 0.400 & 0.425 & 0.032 \\
\bottomrule
\end{tabular*}
\end{table}

\begin{table}[t]
\suppsmall
\begin{minipage}[t]{.56\linewidth}\centering\setlength{\tabcolsep}{2.5pt}
\caption{Every training seed on all test texts. Lens is lens success, drawing is ordinary drawing success per text, and text is the score on the untrained question forms. A dash denotes a missing ordinary-drawing evaluation.}
\label{tab:supp-train-seeds}
\begin{tabular*}{\linewidth}{@{\extracolsep{\fill}}lrrrrrrr@{}}
\toprule
& & \multicolumn{3}{c}{\bagel{}} & \multicolumn{3}{c}{\sensenova{}} \\
\cmidrule(lr){3-5}\cmidrule(l){6-8}
Supervision & Seed & Lens & Drawing & Text & Lens & Drawing & Text \\
\midrule
None & -- & 0.017 & 0.379 & 0.758 & 0.108 & 0.421 & 0.817 \\
\midrule
Text & 0 & 0.025 & 0.396 & 0.917 & 0.142 & 0.388 & 0.913 \\
Text & 1 & 0.017 & 0.412 & 0.908 & 0.133 & 0.379 & 0.896 \\
Text & 2 & 0.025 & 0.425 & 0.904 & 0.192 & 0.396 & 0.921 \\
\midrule
Lens tuning & 0 & 0.892 & 0.508 & 0.772 & 0.975 & 0.521 & 0.736 \\
Lens tuning & 1 & 0.867 & 0.479 & 0.771 & 0.917 & 0.567 & 0.742 \\
Lens tuning & 2 & 0.825 & 0.487 & 0.750 & 0.933 & 0.525 & 0.850 \\
\midrule
Ordinary drawing & 0 & 0.067 & 0.942 & 0.867 & 0.292 & 0.983 & 0.974 \\
Ordinary drawing & 1 & 0.075 & 0.904 & 0.857 & 0.400 & 0.971 & 0.951 \\
Ordinary drawing & 2 & 0.050 & 0.933 & 0.829 & 0.642 & 0.958 & 0.982 \\
\midrule
Word state & 0 & 0.692 & 0.496 & 0.581 & 0.850 & 0.496 & 0.494 \\
Word state & 1 & 0.683 & 0.487 & 0.582 & 0.850 & 0.467 & 0.404 \\
Word state & 2 & 0.658 & 0.471 & 0.576 & 0.858 & 0.517 & 0.546 \\
\midrule
Wrong pictures & 0 & 0.000 & 0.371 & 0.624 & 0.000 & 0.429 & 0.697 \\
Wrong pictures & 1 & 0.000 & -- & 0.590 & 0.000 & 0.421 & 0.633 \\
Wrong pictures & 2 & 0.000 & -- & 0.586 & 0.000 & 0.425 & 0.689 \\
\bottomrule
\end{tabular*}
\end{minipage}\hfill
\begin{minipage}[t]{.42\linewidth}\centering\setlength{\tabcolsep}{3pt}
\caption{Paired differences between kinds of training on the three test sets, by model and measure. Lens success is measured per text pair, and the untrained text and ordinary drawing scores per text. The intervals resample whole noun pairs along with their texts and training seeds.}
\label{tab:supp-train-pairs}
\begin{tabular*}{\linewidth}{@{\extracolsep{\fill}}ll@{}}
\toprule
Comparison & Difference, 95\% interval \\
\midrule
\multicolumn{2}{@{}l}{\bagel{}, lens pairs} \\
Lens minus none & +0.844 [+0.783, +0.900] \\
Lens minus word state & +0.183 [+0.083, +0.289] \\
Text minus none & +0.006 [$-$0.008, +0.025] \\
Ordinary minus none & +0.047 [+0.014, +0.094] \\
\multicolumn{2}{@{}l}{\bagel{}, untrained text} \\
Lens minus none & +0.006 [$-$0.028, +0.038] \\
\multicolumn{2}{@{}l}{\bagel{}, ordinary drawing} \\
Ordinary minus lens & +0.435 [+0.386, +0.483] \\
\midrule
\multicolumn{2}{@{}l}{\sensenova{}, lens pairs} \\
Lens minus none & +0.833 [+0.764, +0.894] \\
Lens minus word state & +0.089 [+0.017, +0.169] \\
Text minus none & +0.047 [$-$0.011, +0.106] \\
Ordinary minus none & +0.336 [+0.247, +0.428] \\
\multicolumn{2}{@{}l}{\sensenova{}, untrained text} \\
Lens minus none & $-$0.041 [$-$0.078, $-$0.002] \\
\multicolumn{2}{@{}l}{\sensenova{}, ordinary drawing} \\
Ordinary minus lens & +0.433 [+0.397, +0.469] \\
\bottomrule
\end{tabular*}
\end{minipage}
\end{table}

The texts test pronoun reference between two objects of different categories, and the judges choose the specified referent, the other named object, both, or neither. Held-out noun labels never occur among the training labels, although familiar appearances can still occur in generated targets. The wrong-picture control is scored for the intended referent. Its zero shows that, among the arms trained with pictures, the referent becomes readable from the pronoun's hidden state only when the pictures depict it.

\subsection{General capabilities after training}
\label{supp:capabilities}
For each model, we compare training seed 0 of text supervision, ordinary drawing supervision, and lens tuning with the untrained model, keeping adapters active throughout evaluation, including drawing. For image generation, we randomly select 20 prompts from each of GenEval's six categories~\citep{ghosh2023geneval} before evaluating any checkpoint and draw one picture per prompt using each model's ordinary text-to-image path with its default settings and drawing seed 0. Our three judges replace GenEval's object detector and answer yes or no for each required object, count, color, and relative position. A prompt passes when at least two judges approve all its requirements. For text knowledge, we use 300 MMLU-Pro test questions~\citep{wang2024mmlupro} drawn at random with seed 0 and select the option whose letter is the most likely first reply token. Our rule written before evaluation defines a skill as harmed when the 95\% paired bootstrap interval for its change lies entirely below zero. By this rule, lens tuning harms neither measure in any model, and all six of its intervals include zero (\cref{tab:supp-capabilities}). Ordinary drawing supervision, in contrast, lowers GenEval by 23.3 points in \bagel{}, mostly on prompts with two objects or a relative position.
\begin{table}[t]
\centering\suppsmall
\caption{General capabilities after training, using training seed 0. Image scores are mean pass rates (\%) across six GenEval prompt categories, with 20 prompts each and our three-judge scoring. Text scores are MMLU-Pro option-letter accuracy (\%) on 300 questions. Changes are percentage points relative to the untrained model, with 95\% paired bootstrap intervals from 10,000 resamples of prompts within categories or of questions.}
\label{tab:supp-capabilities}
\begin{tabular*}{\linewidth}{@{\extracolsep{\fill}}llrlrl@{}}
\toprule
& & \multicolumn{2}{c}{GenEval, image generation} & \multicolumn{2}{c}{MMLU-Pro, text knowledge} \\
\cmidrule(lr){3-4}\cmidrule(l){5-6}
Model & Training & Pass rate & Change, 95\% interval & Accuracy & Change, 95\% interval \\
\midrule
\bagel{} & None & 87.5 & & 28.7 & \\
 & Text & 84.2 & $-$3.3 [$-$6.7, $-$0.8] & 35.0 & +6.3 [+3.3, +9.3] \\
 & Ordinary drawing & 64.2 & $-$23.3 [$-$30.8, $-$15.8] & 26.3 & $-$2.3 [$-$4.7, $-$0.3] \\
 & Lens tuning & 87.5 & +0.0 [$-$5.0, +5.0] & 29.3 & +0.7 [$-$1.0, +2.3] \\
\midrule
\sensenova{} & None & 95.0 & & 28.3 & \\
 & Text & 95.8 & +0.8 [$-$1.7, +3.3] & 31.0 & +2.7 [$-$0.3, +5.7] \\
 & Ordinary drawing & 90.8 & $-$4.2 [$-$9.2, +0.0] & 29.3 & +1.0 [$-$1.7, +4.0] \\
 & Lens tuning & 90.8 & $-$4.2 [$-$9.2, +0.8] & 31.3 & +3.0 [$-$0.7, +7.0] \\
\midrule
\hunyuan{} & None & 95.0 & & 49.7 & \\
 & Text & 95.0 & +0.0 [$-$2.5, +2.5] & 49.0 & $-$0.7 [$-$4.0, +2.7] \\
 & Ordinary drawing & 88.3 & $-$6.7 [$-$11.7, $-$1.7] & 46.7 & $-$3.0 [$-$5.7, $-$0.3] \\
 & Lens tuning & 96.7 & +1.7 [$-$1.7, +5.8] & 47.3 & $-$2.3 [$-$5.7, +0.7] \\
\bottomrule
\end{tabular*}
\end{table}
\subsection{State questions and answer changes}
\label{supp:state-training}

Lens tuning also changes answers to questions about an object's state after an event. Each scenario has two stories, one in which the change happens and one in which it is prevented, followed by a question with two answer options asked in both orders. Both models use the same 40 scenarios, giving 80 stories. The object is named in the question, as in \emph{Is the slice of bread now white or brown?} Twenty of the scenarios were added to an earlier set under the same writing rules and selected from screened candidates using a rule fixed before training. A second set of 28 scenarios (56 stories) uses questions about \emph{it} for the interchange test of \cref{supp:state-interchange}. We supervise the question's subject at the insertion layer in the model's answer format.

The \hunyuan{} checkpoint has 80B parameters, 13B active parameters, and 32 transformer layers shared by text generation and image denoising. We use insertion layer 9, selected by the concrete-word calibration of \cref{supp:implementation}. For text generation, we use the system instruction \emph{Answer every question with exactly one word. Never explain your answer.} Rank-8 LoRA adapters with scaling 16 modify layers 0 to 26 in \showo{} and layers 0 to 8 in \hunyuan{}. They act on the attention and feed-forward projections in \showo{} and on the attention and shared MLP projections in \hunyuan{}. The adapters act only while reading the source text, and the drawing request and image generator use the frozen weights.

Each target is the frozen model's own drawing of \emph{a photo of a \{state\}}, with four target pictures per state in \showo{} and two in \hunyuan{}. The right-state objective uses the state implied by the story, and the wrong-state objective uses the alternative state of the same scenario. Both apply the denoising loss defined in the main paper through the subject's hidden state carried by the \vlens{}. AdamW uses learning rate $10^{-4}$, no weight decay, and batches of eight texts. We train both models for 600 updates with each target. Every setting uses training seeds 0, 1, and 2. Each loss evaluation samples a target picture, fresh noise, and a noise level uniformly on $[0.4,1]$. \hunyuan{} uses noised image latents with timestep $1000t$ and target $\epsilon-z$, and its target pictures are generated at $1024\times1024$ pixels.

A story is answered correctly when the greedy one-word replies to the question in both option orders name the right option, and \cref{tab:supp-state-training} also reports each order separately. We ask every question in both orders because the models prefer one option position. Untrained \showo{} answers 48 of the 80 stories correctly as written and 53 with the options reversed, so a single order mixes the story's effect with this preference. Requiring both orders, or averaging the answer margin over both orders as \cref{fig:states} does, removes it. In the state-question experiment, the pictures of the subject's hidden state use drawing seed 0 at the insertion layer. A picture is correct when two of the three judges choose the right state in the main-subject question of \cref{supp:judging}, whose options are the scenario's two states, both, and neither. The figure's two-option probability is $1/(1+\exp(-m))$, where $m$ is the preferred option's log-probability margin at the first reply token, averaged over both option orders.

\begin{table}[!ht]
\centering\suppsmall
\caption{State-question results for every training seed on the same 80 stories for both models, with lens tuning toward the right or wrong state. The counts show stories with correct one-word replies as written, reversed, and in both orders, and stories pictured correctly.}
\label{tab:supp-state-training}
\begin{tabular*}{\linewidth}{@{\extracolsep{\fill}}llrrrrr@{}}
\toprule
Model & Picture targets & Seed & As written & Reversed & Both orders & Right pictures \\
\midrule
\showo{} & Untrained & -- & 48 & 53 & 29 & 42 \\
\showo{} & Right state & 0 & 55 & 61 & 38 & 59 \\
\showo{} & Right state & 1 & 61 & 50 & 33 & 57 \\
\showo{} & Right state & 2 & 68 & 58 & 48 & 60 \\
\showo{} & Wrong state & 0 & 45 & 54 & 27 & 22 \\
\showo{} & Wrong state & 1 & 46 & 55 & 28 & 23 \\
\showo{} & Wrong state & 2 & 45 & 56 & 28 & 20 \\
\midrule
\hunyuan{} & Untrained & -- & 41 & 38 & 32 & 30 \\
\hunyuan{} & Right state & 0 & 54 & 51 & 45 & 49 \\
\hunyuan{} & Right state & 1 & 47 & 45 & 41 & 55 \\
\hunyuan{} & Right state & 2 & 55 & 46 & 36 & 51 \\
\hunyuan{} & Wrong state & 0 & 52 & 43 & 36 & 37 \\
\hunyuan{} & Wrong state & 1 & 45 & 43 & 36 & 39 \\
\hunyuan{} & Wrong state & 2 & 48 & 47 & 42 & 34 \\
\bottomrule
\end{tabular*}
\end{table}

Right-state pictures raise \showo{}'s correct answers, while wrong-state pictures leave them near the untrained level. Lens tuning with right-state pictures gives 33 to 48 stories answered correctly in both orders, and wrong-state pictures give 27 to 28, against 29 before training (\cref{tab:supp-state-training}). The pictures after wrong-state tuning show the assigned alternative in 53, 49, and 56 of the 80 stories across the three training seeds, compared with 33 before training. In \hunyuan{}, right-state lens tuning gives 36 to 45 stories answered correctly in both orders and wrong-state tuning gives 36 to 42, against 32 before training, so training raises its answers whichever state the pictures show. Its pictures after wrong-state tuning show the assigned alternative in 34, 37, and 37 of the 80 stories, compared with 24 before training. In \showo{}, the avocado, butter, caterpillar, hen's egg, and pine stories change from a wrong reply to the right one in both option orders and show the right state in all three right-state runs, and four more stories do so in two. In \hunyuan{}, the marshmallow, pizza, and sandcastle stories do so in all three runs, and nine more do so in two.

\subsection{Interchanging the subject's hidden states}
\label{supp:state-interchange}

For the second set of 56 stories, we carry the trained subject word's hidden states from each of the three \showo{} donors lens-tuned on these stories into the untrained model while it answers the identical question. At each layer from 0 to 27, the subject's hidden state is replaced by its raw state from the trained model's reading of that text. All other token positions are recomputed by the untrained model while no adapters are active. We use both option orders and repeat the intervention for all three trained models. Self-swapping each state reproduces every answer margin exactly in both the original and trained models.

The trained subject's hidden states raise the untrained model's correct answers from 41 to 47, 47, and 49 of 56, carrying 38\%, 39\%, and 21\% of the mean-margin gain (\cref{tab:supp-state-interchange}). The transferred share is the interchange margin minus the original margin, divided by the trained margin minus the original margin. Replacing only the state entering layer 27 leaves the mean margin within 0.01 of the original value. Transferring the subject's states across layers carries part of the answer change, though this swap does not isolate the information shown in their pictures.

\begin{table}[!ht]
\centering\suppsmall
\caption{The trained subject's hidden states carry answer corrections into the untrained \showo{}. The original model answers 41/56 stories correctly with mean margin +1.023, and the three trained donors answer 47, 46, and 51 with mean margins +1.97, +2.07, and +3.50. Each row takes states from one trained model and keeps the recipient's weights fixed. The carried share compares the margin gain from interchange with the gain from that trained model. Carried shares use unrounded mean margins.}
\label{tab:supp-state-interchange}
\begin{tabular*}{\linewidth}{@{\extracolsep{\fill}}lrrrrr@{}}
\toprule
& \multicolumn{3}{c}{Subject states interchanged at all layers} & \multicolumn{2}{c}{At layer 27 only} \\
\cmidrule(lr){2-4}\cmidrule(lr){5-6}
Training seed & Answers & Mean margin & Margin gain carried & Answers & Mean margin \\
\midrule
0 & 47 & +1.38 & 38\% & 40 & +1.032 \\
1 & 47 & +1.43 & 39\% & 40 & +1.029 \\
2 & 49 & +1.53 & 21\% & 40 & +1.019 \\
\bottomrule
\end{tabular*}
\end{table}

\section{How the items were built}
\label{supp:items}
All items were drafted and checked as text before any picture was drawn, and the exact texts are stored with the code. \Cref{tab:supp-examples} shows one example from each of the item sets described here. The matched descriptions (30 pairs) are questions whose description of an object ends in a part word shared by both objects of the pair. Neither object is named, and each question asks for a short factual answer. The two objects are distinct main subjects rather than a part and its whole. The pronoun pairs (20 pairs) name two things and then select one with a verb. Ten pairs change the verb while keeping both things, and ten keep the verb while changing one of the things. The items with two meanings (20 words) use one word twice in a sentence, with the selecting context before each occurrence, in both orders. The 16 depth-study questions each include a preceding sentence that names the object and an alternative object for the judges. The two-state items pair objects that share two color or pattern values, with one two-sentence text per value that mentions the object and later refers to it with \emph{it}, and one plain sentence per value that names both the object and the value.

\begin{table}[!t]
\centering\suppsmall
\caption{One example from each item set of the context, depth, two-state, and lens tuning studies, with the word whose hidden state is read and the object the picture should show. The state stories appear in \cref{sec:state-training} and the held thoughts in \cref{supp:held}.}
\label{tab:supp-examples}
\begin{tabular*}{\linewidth}{@{\extracolsep{\fill}}P{.17\linewidth}P{.54\linewidth}P{.08\linewidth}P{.14\linewidth}@{}}
\toprule
Study & Example texts & Read & Intended \\
\midrule
A, matched descriptions, 30 pairs & How many legs does the huge land animal with a long trunk and tusks have? \newline Does the whiskered Arctic animal with flippers and long tusks breathe air? & tusks & elephant \newline walrus \\
\midrule
B, reference, verb changed, 10 pairs & A sheep and a bicycle stood in the yard. She sheared it. \newline A sheep and a bicycle stood in the yard. She pedaled it. & it & sheep \newline bicycle \\
\midrule
B, reference, verb held, 10 pairs & A dog and a teapot were in the yard. She fed it. \newline A horse and a teapot were in the yard. She fed it. & it & dog \newline horse \\
\midrule
C, two meanings, 20 words in both orders & Hanging upside down in the attic, a furry bat slept; among the baseball equipment lay a wooden bat. \newline Among the baseball equipment lay a wooden bat; hanging upside down in the attic, a furry bat slept. & bat, bat & bat animal \newline baseball bat \\
\midrule
Depth study, 16 questions & How many legs does the animal that spins webs have? \newline With the name given first.  \emph{A spider is an animal that spins webs.} How many legs does the animal that spins webs have? & webs & spider, alternative stop sign, answer eight or 8 \\
\midrule
Two states, 8 development pairs & The car was painted bright red. Later I looked at it. \newline Plain sentence: I saw a red car on the street. \newline Partner: bicycle, values red and blue & red, it & red car \\
\midrule
Two states, 8 held-out pairs & Someone painted the bus red. I drove it home. \newline Plain sentence: There was a red bus near the station. \newline Partner: kite, values red and green & red, it & red bus \\
\midrule
Training texts, 960 in the pool & We found a cake near a radio. We ate it. \newline Sentence forms and nouns are listed in \cref{tab:supp-nouns}. & it & cake \\
\bottomrule
\end{tabular*}
\end{table}

For the training study, a fixed script with seed 20261019 combines the nouns, selecting expressions, and category questions in \cref{tab:supp-nouns}. Animals and food can pair with any other category, devices and openable objects pair only with animals or food, and vehicles additionally pair with clothing, so that the selecting verb never fits both objects. The script balances noun frequency and prevents repeats between training and tests when selecting pairs. The naming question asks what \emph{it} refers to in the last sentence, the untrained choice asks whether \emph{it} is X or Y with the order of X and Y randomized, and the two yes or no questions ask the category question for the referent and for the alternative. The four neutral sentences for the word-state targets are \emph{There is a X here.}, \emph{I looked at the X.}, \emph{This is a picture of a X.}, and \emph{She talked about the X for a while.}

\begin{table}[!t]
\centering\suppsmall
\caption{The 84 nouns of the training study by category and split, each category's selecting expressions and yes or no question, and the six sentence forms, four for training and two held out. X and Y are nouns, VERB includes the pronoun, and PLACE is kitchen, yard, garage, hall, barn, or shed. Because we read the hidden state at the pronoun, later particles cannot affect it.}
\label{tab:supp-nouns}
\begin{tabular*}{\linewidth}{@{\extracolsep{\fill}}P{.08\linewidth}P{.31\linewidth}P{.12\linewidth}P{.12\linewidth}P{.15\linewidth}P{.13\linewidth}@{}}
\toprule
Category & Training nouns & Development & Held-out & Expressions, training / held-out & Category question \\
\midrule
animal & dog, rabbit, cow, goat, donkey, camel, deer, kangaroo & cat, horse, pig & sheep, hamster, llama & fed it, petted it / groomed it & Is it an animal? \\
food & apple, banana, orange, pear, cake, sandwich, cookie, watermelon & pizza, pineapple, strawberry & carrot, donut, burger & ate it, tasted it / bit into it & Is it food? \\
device & radio, laptop, microwave, vacuum cleaner, hair dryer, printer, washing machine, sewing machine & lamp, toaster, blender & fan, television, kettle & switched it on, turned it off / plugged it in & Does it run on electricity? \\
vehicle & truck, bus, tractor, van, bulldozer, motorcycle, golf cart, snowmobile & taxi, jeep, forklift & car, ambulance, limousine & drove it, parked it / steered it & Is it a vehicle? \\
openable & umbrella, book, backpack, wallet, envelope, briefcase, purse, bottle & box, jar, gate & suitcase, cupboard, refrigerator & opened it, closed it / shut it & Can you open and close it? \\
clothing & jacket, scarf, boot, glove, dress, shirt, skirt, belt & sock, sneaker, apron & hat, sweater, tie & wore it, put it on / took it off & Do people wear it? \\
\midrule
Forms & \multicolumn{5}{@{}P{.90\linewidth}@{}}{\emph{A X and a Y were in the PLACE. She VERB.} \quad \emph{There was a X next to a Y. He VERB.} \quad \emph{I saw a X and a Y in the PLACE. Then I VERB.} \quad \emph{We found a X near a Y. We VERB.}} \\
Held-out & \multicolumn{5}{@{}P{.90\linewidth}@{}}{\emph{A X sat beside a Y all morning. Later my aunt VERB.} \quad \emph{Our neighbour kept a X and a Y. One day he VERB.}} \\
\bottomrule
\end{tabular*}
\end{table}

\section{More pictures}
\label{supp:galleries}
The galleries visualize hidden states with the main paper's lens. \Cref{fig:supp-layers} keeps the token and the insertion layer fixed and varies only the extraction depth. \Cref{fig:supp-unnamed} contrasts a word written into the drawing request with the same word's hidden state read inside a question. \Cref{fig:supp-meanings} reads both occurrences of an ambiguous word from one forward pass.

\begin{figure}[!t]\centering
\includegraphics[width=\linewidth]{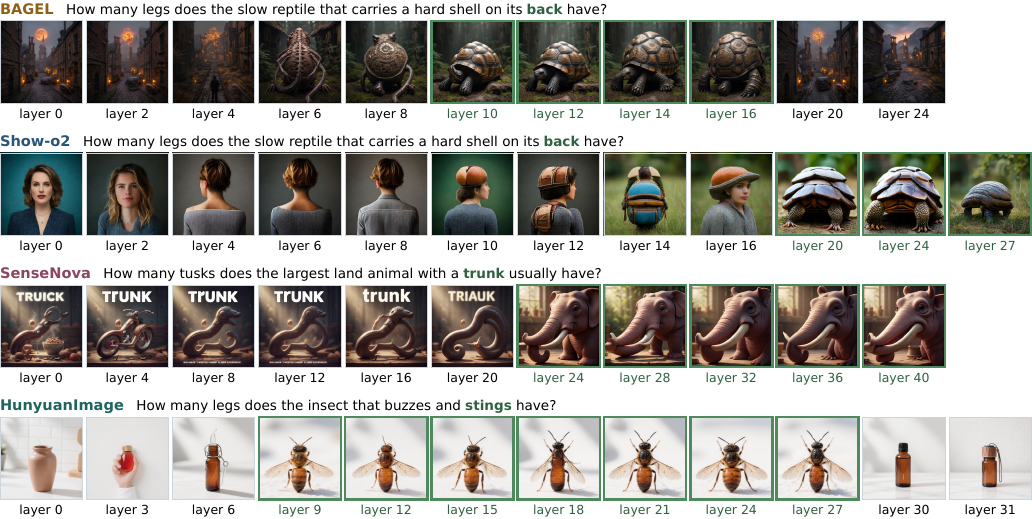}
\caption{One token's picture at selected extraction depths, with the insertion layer fixed. The rows read \emph{back} in the turtle question (\bagel{}, \showo{}), \emph{trunk} in the elephant question (\sensenova{}), and \emph{stings} in the bee question (\hunyuan{}). Framed pictures show the described animal, and the others show unrelated scenes, people, objects, or written words.}
\label{fig:supp-layers}
\end{figure}

\begin{figure}[!t]\centering
\includegraphics[width=\linewidth]{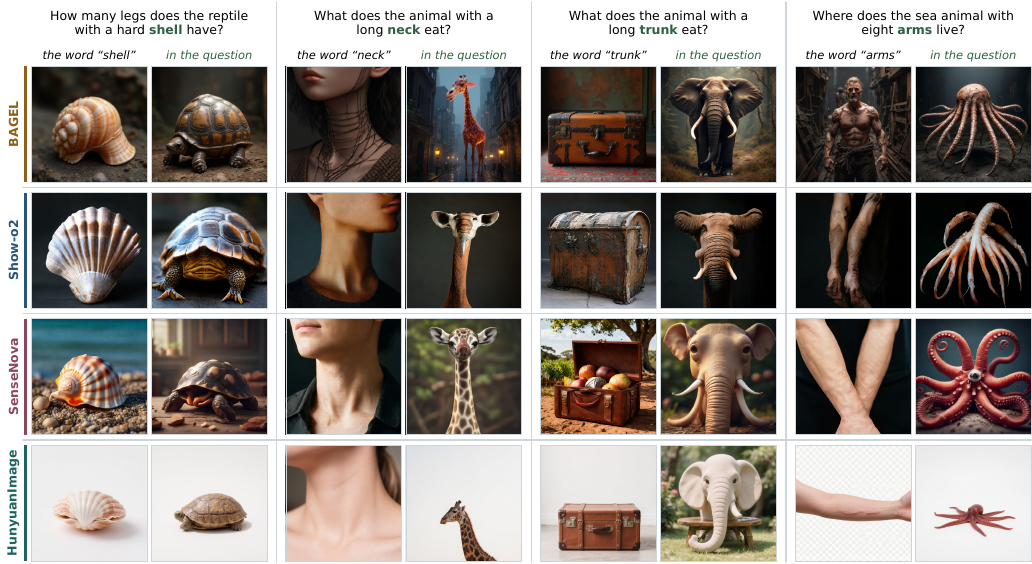}
\caption{Objects that a question describes but never names, read at each model's insertion layer. Each row uses the same four questions. In each pair, the left picture shows the highlighted word in the drawing request. The right picture visualizes its hidden state inside the question.}
\label{fig:supp-unnamed}
\vspace{0.8em}
\includegraphics[width=\linewidth]{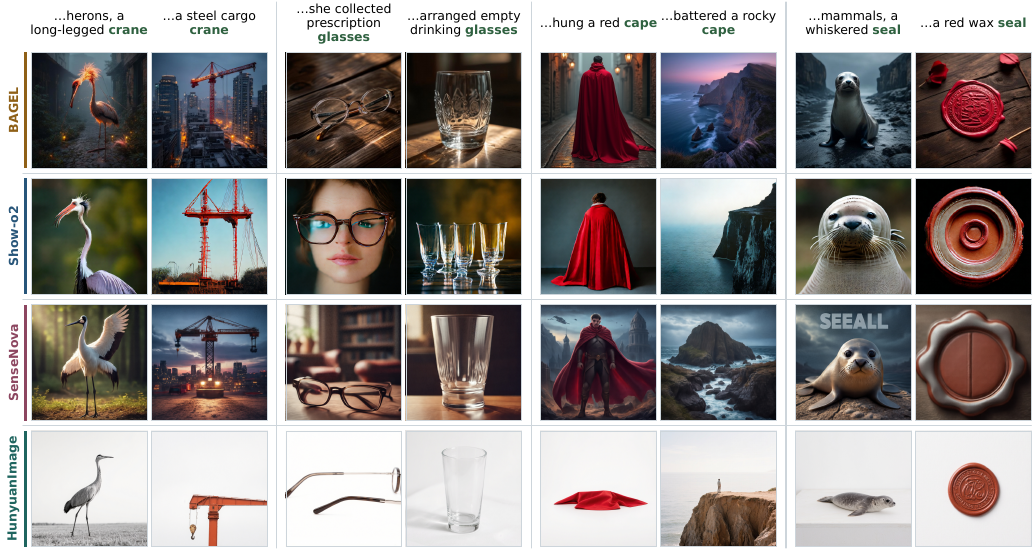}
\caption{Two meanings of one word in one sentence, read at each model's insertion layer. Each row reads the same four sentences of the two-meaning test from one forward pass, and each column reads the occurrence that ends the text above it.}
\label{fig:supp-meanings}
\end{figure}

\clearpage
\section{Autoregressive image generators}
\label{supp:autoregressive}

We use the 7B checkpoints of \liquid{} and \januspro{}. \liquid{} has 28 language-model layers and writes 1,024 discrete VQ image codes at $512\times512$ pixels with classifier-free guidance 7. \januspro{} has 30 language-model layers and writes 576 discrete image codes through a separate generation head at $384\times384$ pixels with guidance 5. The insertion layers $R=6$ and $R=8$ are the latest layers at which CLIP ranks all 12 calibration words of \cref{supp:implementation} first. The scaled hidden state enters \emph{thing} in \emph{a photo of a thing} once, and image tokens read it through the key/value cache. \Cref{tab:supp-autoregressive} reports the two-meaning test of \cref{supp:context} at $R$, with 20 words in both sentence orders, drawing seeds 0, 1, and 2, and the same judges and success rules.

\begin{table}[!ht]
\centering\suppsmall
\caption{Words with two meanings whose two meanings are both visualized at the insertion layer, averaged over the two sentence orders, with the label-swap chance and the one-sided $p$-value.}
\label{tab:supp-autoregressive}
\begin{tabular*}{\linewidth}{@{\extracolsep{\fill}}lrrr@{}}
\toprule
Model & Words right at $R$ & Chance & $p$ \\
\midrule
\liquid{} & 12/20 & 0.300 & 0.0002 \\
\januspro{} & 11.5/20 & 0.287 & 0.0001 \\
\bottomrule
\end{tabular*}
\end{table}

\paragraph{Depth study.}
We repeat the depth study of \cref{supp:depth} on both models with the same 16 questions, token positions, judges, and success rules. We read every second layer and the final layer with drawing seeds 3, 4, and 5. A question counts when the model answers it correctly without naming the object, leaving 13 questions for \liquid{} and 6 for \januspro{}. \Cref{tab:supp-ar-depth} reports every layer, and the \liquid{} curves appear in \cref{fig:layers}. In \liquid{}, the description's last word shows the object in 11/13 questions at $R$ and in 12/13 at layer 8. Later tokens show it in 10 to 11 questions from layer 22 on, mostly at chat-format tokens and the question mark, and the token immediately before the answer shows it in up to 4. In \januspro{}, the description's last word shows the object in 4/6 questions at $R$ and in at most 2/6 at any other layer, while later tokens recover it in at most 1/6.

\begin{table}[!ht]
\centering\suppsmall\setlength{\tabcolsep}{3pt}
\caption{Depth study for the two autoregressive models. The counts show questions whose pictures contain the described object at the description's last word and at the best later token, and objects visualized from their names in plain sentences. Bold marks $R$.}
\label{tab:supp-ar-depth}
\begin{tabular*}{\linewidth}{@{\extracolsep{\fill}}l*{16}{r}@{}}
\toprule
\liquid{} layer & 0 & 2 & 4 & \textbf{6} & 8 & 10 & 12 & 14 & 16 & 18 & 20 & 22 & 24 & 26 & 27 & \\
Description's last word (of 13) & 2 & 7 & 8 & \textbf{11} & 12 & 11 & 11 & 11 & 11 & 11 & 11 & 11 & 11 & 11 & 11 & \\
Best later token (of 13) & 1 & 4 & 4 & \textbf{3} & 8 & 6 & 4 & 3 & 3 & 4 & 5 & 10 & 10 & 11 & 10 & \\
Name in a plain sentence (of 15) & 8 & 15 & 15 & \textbf{15} & 15 & 15 & 15 & 15 & 15 & 15 & 15 & 15 & 15 & 14 & 15 & \\
\midrule
\januspro{} layer & 0 & 2 & 4 & 6 & \textbf{8} & 10 & 12 & 14 & 16 & 18 & 20 & 22 & 24 & 26 & 28 & 29 \\
Description's last word (of 6) & 0 & 0 & 0 & 0 & \textbf{4} & 2 & 1 & 0 & 1 & 0 & 1 & 1 & 0 & 0 & 0 & 0 \\
Best later token (of 6) & 0 & 0 & 0 & 0 & \textbf{0} & 1 & 0 & 0 & 1 & 0 & 0 & 0 & 1 & 0 & 0 & 1 \\
Name in a plain sentence (of 15) & 0 & 8 & 13 & 15 & \textbf{14} & 14 & 14 & 14 & 14 & 14 & 13 & 12 & 9 & 14 & 9 & 6 \\
\bottomrule
\end{tabular*}
\end{table}

\section{Held thoughts while copying}
\label{supp:held}

We extend the examples in \cref{sec:held-thoughts} with a search whose protocol was fixed before drawing any pictures. We ask each model to copy one of 14 neutral sentences. \showo{} is asked to think about one of 12 concrete objects and \hunyuan{} about one of 8, with none of these objects named in the sentences. We retain replies whose similarity to the requested sentence is at least 0.85. We read every token of each retained copy through the \vlens{} at layers 24, 25, and 26 with drawing seeds 0, 1, and 2 (in \hunyuan{} seed 2 only where the first two disagree) and the same three judges. A token-layer pair succeeds when it shows the intended object in at least two seeds. As a sanity check, we read each token at each layer in the same way for the same sentence copied without the think instruction and with a different object requested.

In \showo{}, 97 of the 105 successful token-layer pairs occur at \emph{the}. \hunyuan{} often rewrites the sentence, leaving 46 retained copies from 112 attempts, and 12 of its 13 successful token-layer pairs occur at the final period, mostly at layers 25 and 26. The sanity checks almost never show the requested object. \Cref{fig:supp-held,fig:supp-held-hunyuan} provide examples.

\begin{figure*}[t]
    \centering
    \includegraphics[width=\linewidth]{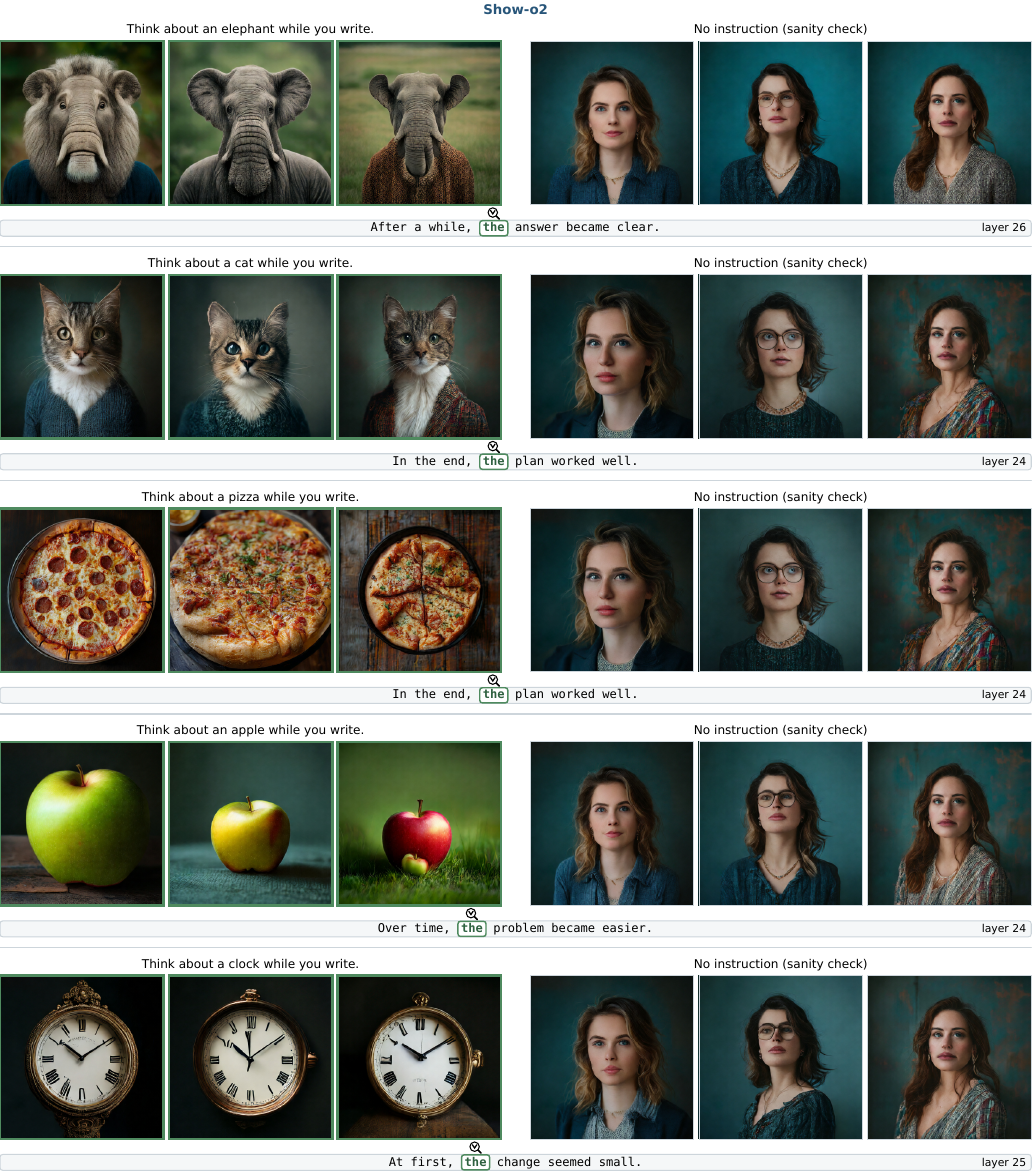}
    \caption{Objects pictured while \showo{} copies an unrelated sentence. Each row shows three pictures generated from the hidden state of \emph{the} in the model's own copy at the indicated layer, with accepted pictures framed in green. The same token read without the think instruction follows as a sanity check. The model's copy appears underneath with the read token marked.}
    \label{fig:supp-held}
\end{figure*}

\begin{figure*}[t]
    \centering
    \includegraphics[width=\linewidth]{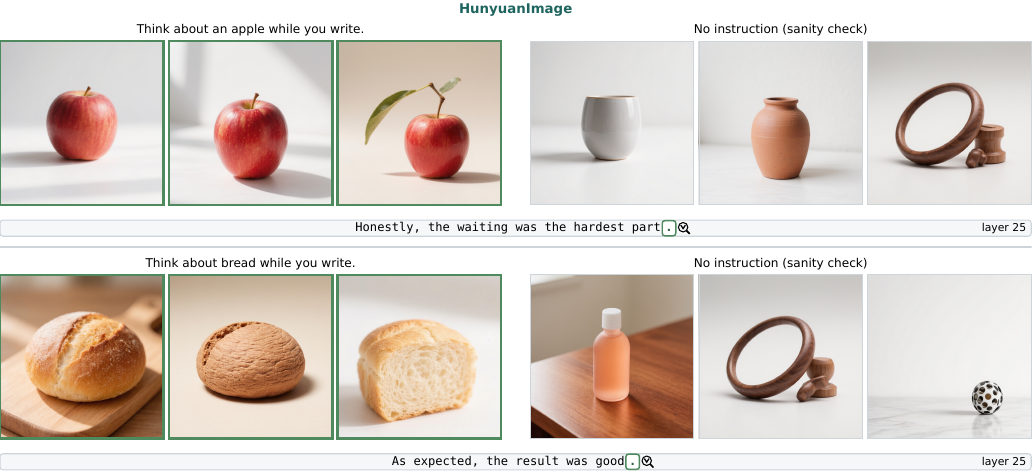}
    \caption{Corresponding examples for \hunyuan{}, read at the final period of its copy. The rows are selected examples, and the displayed drawing seeds were also selected. The sanity check uses the same seeds.}
    \label{fig:supp-held-hunyuan}
\end{figure*}

\section{Reproducing the results}
\label{supp:reproduce}

The code accompanies the supplementary material. Its \texttt{vlens} package contains the model adapters, the \vlens{} transport, and code for estimating its statistics, drawing pictures, lens tuning, judging, and scoring. The release also includes scripts for each experiment, item files, saved test-set selections, and environment requirements for the models and judges. The README explains installation and checkpoint downloads. The script \texttt{third\_party/setup.sh} fetches upstream code at pinned commits and applies small patches, while the environment instructions cover the remaining dependencies. All models and judges use the public checkpoints listed in the README.

Begin by calibrating the insertion layer and estimating the transport statistics from the supplied text passages. Then run the experiment's state extraction and drawing steps, including training where needed, with the seeds stated in the paper. Run all three judges on the resulting pictures and then use the CPU scorers to combine their answers into the reported tables. \Cref{tab:supp-reproduce} maps results to their scripts, and the README provides the commands, model settings, and expected outputs. Drawing jobs can be split across GPUs and resumed without repeating completed pictures.

\begin{table}[t]
\centering\suppsmall
\caption{Scripts for reproducing the paper's results, all in \texttt{scripts/}. Drawing outputs are judged with \texttt{run\_judge.py} or the experiment's own judging step before scoring. The README lists the commands in execution order.}
\label{tab:supp-reproduce}
\begin{tabular*}{\linewidth}{@{\extracolsep{\fill}}P{.34\linewidth}P{.62\linewidth}@{}}
\toprule
Result & Scripts \\
\midrule
Insertion layers (\cref{tab:supp-calibration}, \cref{supp:autoregressive})
& \texttt{calibrate\_insertion\_layer.py} \\

Transport statistics (\cref{eq:transport})
& \texttt{fit\_transport.py}, \texttt{collect\_states.py}, \texttt{build\_lens\_stats.py}, \texttt{lens\_stats\_own\_tokens.py} \\

Lens variants and final normalization (\cref{tab:variants}, \cref{tab:supp-variants,tab:supp-final-norm})
& \texttt{draw\_cells.py}, \texttt{self\_gain.py}, \texttt{score\_cells.py}, \texttt{score\_variant\_choice.py}, \texttt{score\_variants.py} \\

Ablation (\cref{tab:ablation})
& \texttt{lens\_ablation.py}, \texttt{draw\_cells.py}, \texttt{score\_cells.py}, \texttt{score\_ablation.py} \\

Judge calibration (\cref{tab:supp-judge})
& \texttt{draw\_judge\_calibration.py}, \texttt{run\_judge.py}, \texttt{score\_judge\_calibration.py} \\

Context (\cref{tab:context,tab:supp-context})
& \texttt{draw\_cells.py}, \texttt{score\_context.py} \\

Autoregressive models (\cref{sec:autoregressive})
& \texttt{draw\_cells.py}, \texttt{score\_context.py} \\

Tokens and depth (\cref{fig:layers,supp:depth,tab:supp-ar-depth})
& \texttt{depth\_fixed\_positions.py}, \texttt{depth\_all\_tokens.py}, \texttt{depth\_through\_answer.py}, \texttt{depth\_token\_report.py}, \texttt{depth\_autoregressive.py} \\

Text readouts (\cref{tab:supp-text-readouts})
& \texttt{fit\_text\_jlens.py}, \texttt{fit\_tuned\_lens.py}, \texttt{text\_lens\_ranks.py}, \texttt{score\_text\_lenses.py}, \texttt{text\_readout\_depth.py} \\

Held thoughts (\cref{sec:held-thoughts,supp:held})
& \texttt{held\_thoughts.py}, \texttt{held\_search.py} \\

Two states (\cref{supp:two})
& \texttt{two\_states.py}, \texttt{two\_states\_presence.py} \\

Pronoun tuning (\cref{tab:supervision})
& \texttt{make\_tuning\_items.py}, \texttt{lens\_tuning.py}, \texttt{score\_lens\_tuning.py} \\

General capabilities (\cref{tab:supp-capabilities})
& \texttt{benchmarks.py}, \texttt{capability.py} \\

State questions (\cref{tab:states})
& \texttt{make\_state\_items.py}, \texttt{state\_tuning.py}, \texttt{state\_table.py} \\

State interchange (\cref{supp:state-interchange})
& \texttt{state\_tuning.py} with \texttt{swap} and \texttt{swap\_table} \\
\bottomrule
\end{tabular*}
\end{table}

Jobs use one A100-80GB or L40S-48GB GPU for \bagel{}, \showo{}, \sensenova{}, \liquid{}, and \januspro{}, and four A100-80GB or six L40S-48GB GPUs for \hunyuan{}. The larger runs require several hundred GPU hours for lens variants, about 480 for ablation drawings plus 110 for judging, and about 210 per \hunyuan{} pronoun tuning seed. Generated pictures, fitted statistics, judge answers, and model weights are not bundled with the release. The human rating study in \cref{tab:supp-human} is also excluded because the raters' answers are personal data.

\section{Limitations}
\label{supp:limitations}

The \vlens{} depends on what the model's generator can draw. We study objects, colors, and visible states, and do not establish reliable readings of numbers, relations, or abstract ideas. Each reading also requires generating a full image, taking 9--110 seconds in our setup. A missing picture does not establish that a hidden state lacks the concept, as the untrained pronoun results illustrate. Our scores depend on three multimodal judges choosing among a few options, and calibration and human checks cannot rule out acceptance of similar but incorrect objects. The item sets are small and mostly written by hand, with 16 two-hop questions, 30 description pairs, 20 two-meaning words, 20 pronoun pairs, 16 two-state pairs, 40 state scenarios, and 28 interchange scenarios, in addition to the lens-variant sets of \cref{sec:supp-variants}. Recovery of unnamed objects is modest and is not significantly above chance in \bagel{} or \sensenova{}. The best later token in \cref{fig:layers} selects among many positions for each question and layer, so its curve is a descriptive summary.

Lens tuning trains hidden states through the same lens used to evaluate them. Gains in ordinary drawing are small, and word state supervision without pictures exceeds lens tuning on unseen phrases in \sensenova{} and on unseen nouns and phrases in \hunyuan{}. The gains in text answers correspond to about ten stories out of 80 and are measured on the supervised stories. The interchange test transfers whole hidden states and has no control donor trained on different picture targets, so it does not isolate the contribution of the pictured content. Finally, the general capability checks use one training seed, and their intervals still allow losses of several percentage points.

\clearpage

\fi
\end{document}